\documentclass[journal]{style/IEEEtran}

\IEEEoverridecommandlockouts                              
\makeatletter

\let\proof\@undefined
\let\endproof\@undefined

\let\NAT@parse\undefined
\makeatother

\usepackage[numbers, sort, compress]{natbib}
\usepackage{graphicx}
\usepackage{caption}
\usepackage{amsmath,amssymb}
\usepackage{amsthm} 
\usepackage{booktabs} 
\usepackage[usenames,dvipsnames,svgnames,table]{xcolor}
\usepackage{subfig} 
\usepackage{placeins}
\usepackage{algpseudocode}

\usepackage{multicol}
\usepackage{algorithm}

\usepackage{balance}
\usepackage{mathrsfs}

\usepackage{verbatim}
\usepackage{float}

\usepackage{style/perl_acronyms}
\usepackage{style/perl_math}
\usepackage{style/perl_SIunits}
\usepackage{style/perl_misc}

\usepackage[inline]{enumitem}

\newcommand{\xib}[1]{\boldsymbol{\xi}_{#1}}
\newcommand{\xip}[1]{\boldsymbol{\xi}^\prime_{#1}}
\newcommand{\xic}[1]{\boldsymbol{\xi}^\curlywedge_{#1}}
\newcommand{\xipc}[1]{\boldsymbol{\xi}^{\prime\curlywedge}_{#1}}
\newcommand{\xibt}[1]{\boldsymbol{\xi}^\top_{#1}}
\newcommand{\xipt}[1]{\boldsymbol{\xi}^{\prime\top}_{#1}}
\newcommand{\xict}[1]{\boldsymbol{\xi}^{\curlywedge\top}_{#1}}
\newcommand{\xipct}[1]{\boldsymbol{\xi}^{\prime\curlywedge\top}_{#1}}

\newtheoremstyle{nospace}{2pt}{1.5pt}{\itshape}{}{\bfseries}{:}{.5em}{}
\theoremstyle{nospace}

\begin{document}
\title{Characterizing the Uncertainty of Jointly Distributed Poses in the Lie~Algebra}

\author{Joshua~G.~Mangelson, ~Maani~Ghaffari, ~Ram~Vasudevan,~and~Ryan~M.~Eustice%
  \thanks{*This work was supported by the \acl{ONR} under awards
    N00014-16-1-2102. Funding for M. Ghaffari is given by the Toyota Research Institute (TRI), partly under award number N021515, however this article solely reflects the opinions and conclusions of its authors and not TRI or any other Toyota entity.}%
  \thanks{J.~Mangelson, M.~Ghaffari, R.~Vasudevan, and R.~Eustice are at the University of Michigan, Ann Arbor, MI 48109, USA. \texttt{\{mangelso, maanigj, ramv, eustice\}@umich.edu}.}  %
}


\maketitle

\begin{abstract}
An accurate characterization of pose uncertainty is essential for safe
autonomous navigation. Early pose uncertainty characterization methods
proposed by Smith, Self, and Cheeseman (SCC), used coordinate-based
first-order methods to propagate uncertainty through non-linear
functions such as pose composition (head-to-tail), pose inversion, and
relative pose extraction (tail-to-tail). Characterizing uncertainty in
the Lie Algebra of the special Euclidean group results in better
uncertainty estimates. However, existing approaches assume that
individual poses are independent. Since factors in a pose graph induce
correlation, this independence assumption is usually not reflected in reality. In addition, prior work has focused primarily on the pose composition operation. This paper develops a framework for modeling the uncertainty of jointly distributed poses and describes how to perform the equivalent of the SSC pose operations while characterizing uncertainty in the Lie Algebra. Evaluation on simulated and open-source datasets shows that the proposed methods result in more accurate uncertainty estimates. An accompanying C++ library implementation is also released.
\end{abstract}

\begin{IEEEkeywords}
SLAM, mobile robotics, uncertainty propagation, Lie group, Lie algebra, matrix groups, rigid body transformation, state estimation. 
\end{IEEEkeywords}
\acresetall

\IEEEpeerreviewmaketitle

\begin{figure}[t!]
    \centering
    \includegraphics[width=0.9\columnwidth]{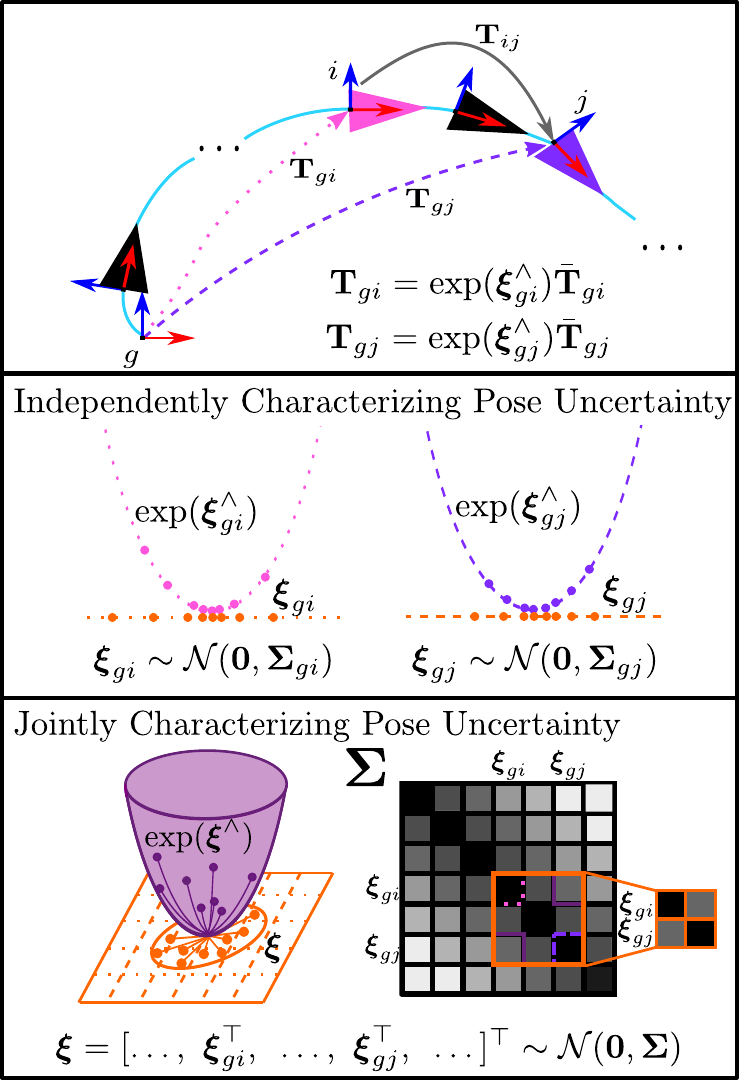}
    \caption{
    State-of-the-art Pose-Graph SLAM algorithms estimate the pose (depicted as $i$ and $j$ in the top illustration) of a robotic vehicle at each time step with respect to a fixed coordinate frame, $g$, which is denoted by $\mathbf{T}_{gi}$ and $\mathbf{T}_{gj}$, respectively. 
    After solving SLAM, it is often necessary to extract additional information by performing a variety of operations such as pose composition, pose inversion, and relative pose estimation, while accurately propagating uncertainty.
    An example of the relative pose operation $\mathbf{T}_{ij}$ is shown at the top of this figure.
    Recent work has shown that characterizing uncertainty as Gaussian random variables ($\boldsymbol{\xi}_{gi}, \boldsymbol{\xi}_{gj}$) in the Lie algebra of the Special Euclidean group (shown in the middle of the above figure) leads to increased consistency \cite{barfoot2014associating}; however, this approach has focused on pose composition while assuming that the underlying poses are independent.
    Typically, the poses estimated from SLAM are heavily correlated \cite{dissanayake2001a}.
    This paper proposes a framework for jointly characterizing the uncertainty of a set of correlated poses in the Lie algebra space (shown in the bottom illustration of the above figure). 
    It then describes how to perform the pose composition, pose inverse, and relative pose operations within this framework. }
    \label{fig:main_example}
\end{figure}

\section{Introduction}

\IEEEPARstart{A}{n} accurate characterization of robot pose (location and orientation) uncertainty is essential to robust
long-term autonomy because planning and safety decisions are often predicated on their value \cite{thrun2005probabilistic}. For example, an 
over-confident position estimate could potentially result in a self-driving car crossing out of its lane or an underwater
vehicle colliding with a submerged structure. On the other hand, under-confidence can lead to slow or sluggish behavior.  

One of the first papers to characterize pose uncertainty of coordinate frame relationships represents the relative pose of objects using a multivariate Gaussian parameter vector and associated covariance matrix \cite{smith1986a}. 
This paper was later extended by Smith, Self, and Cheesman \cite{smith1990a} by representing multiple uncertain spatial relationships as a \textit{stochastic map} which could be used to evaluate the uncertainty of any given pose with respect to any other. 
They also proposed several operations (such as the relative pose operation shown in \figref{fig:main_example}) that enable the extraction of additional information not directly estimated, along with first order coordinate-based methods for propagating uncertainty through these operations. 
For brevity, the operations proposed in \cite{smith1990a} are often referred to by the initials of the paper's authors (SSC). 

Although it is well-understood that the rigid body transformation (or the motion group of $\mathbb{R}^3$) is described by the three dimensional (3D) Special Euclidean group~\citep{spong2005robot,murray1994mathematical}, $\mathrm{SE}(3)$, the uncertainty of these transformations is often modeled in local coordinates leading to inconsistencies in the estimation problem~\citep{huang2007convergence} or the loss of monotonicity in uncertainty propagation~\citep{rodriguez2018importance}. 
 \citet{wang2008nonparametric} and \citet{long2013banana} were able to overcome these problems by 
 representing each pose using \emph{exponential coordinates} located in the Lie algebra of the
 $\mathrm{SE}(d)$. 
 
\citet{barfoot2014associating} were then able to show that propagation computations could be simplified by modeling the uncertainty directly in the Lie algebra and then using the exponential map to induce
a distribution in the group space. 
Since the Lie algebra is a vector space, a small perturbation term can be modeled as zero-mean Gaussian noise in $\mathbb{R}^6$ and then used to perturb a mean (or nominal) pose in the group space.
\cite{barfoot2014associating} then builds on this by deriving first and second order uncertainty 
propagation for the pose composition operation when the associated poses are independent. 
Our approach for modeling the uncertainty of a set of poses is similar, however we drop the independence requirement since the poses estimated by SLAM are rarely independent and additionally describe the additional operations of pose inversion and relative pose extraction (See \figref{fig:main_example}).

The main contributions of this paper are as follows:
\begin{enumerate}
    \item we present a framework that describes how to represent jointly correlated poses while using the Lie algebra to characterize uncertainty;
    \item we derive the equivalent of the SSC operations under the proposed framework;
    \item we describe how to convert from alternative uncertainty characterization parameterizations to the proposed framework  (including Lie algebra covariance extraction from a MLE solution); and,
    \item we release an accompanying C++ library implementation along with examples presented here. 
\end{enumerate}
The remainder of this paper is organized as follows:
Section \ref{sec:SE3} provides a brief introduction to the Special Euclidean group and some necessary concepts from Lie group Theory. 
Section \ref{sec:SSC} provides a summary of the SSC uncertainty representation framework and its associated operations. 
Section \ref{sec:lie_joint_uncertainty} describes how to use the Lie algebra to characterize uncertainty for jointly distributed poses. 
Sections \ref{sec:pose_composition}, \secref{sec:inverse}, and \secref{sec:relative_pose} describe the derivation of the pose composition, pose inversion, and relative pose operations, respectively, while characterizing uncertainty on the Lie algebra.
Section \ref{sec:conversion} describes how to convert from a coordinate based representation of uncertainty to the Lie algebra based representation and how to extract an estimate of pose uncertainty from a MLE solution.
Section \ref{sec:eval} describes an experimental evaluation of the proposed methods.
Section \ref{sec:library} describes the implementation of the released library. 
Finally, Section \ref{sec:conclusion} concludes the paper. 

\section{The Special Euclidean Group and \\Lie Group Theory}
\label{sec:SE3}

Estimation of the relative pose (position and orientation) between objects or coordinate frames
in space is a common problem in robotic navigation, perception, and manipulation. 
Formally, we represent 3D relative pose transformations as elements of the \textit{Special Euclidean group}. 
This section provides a brief introduction to the \textit{Special Euclidean group} and relevant aspects of Lie group theory that are important during the subsequent derivation and discussion.  

\subsection{The Special Euclidean group}

The \textit{Special Euclidean group}, or $\mathrm{SE}(d)$, represents the space of homogeneous transformation matrices or the space of matrices that apply a rigid body rotation and translation to points in $\mathbb{R}^d$ (represented in homogeneous form).
Formally, in three dimensions, $\mathrm{SE}(3)$ is defined as follows:
\begin{equation}
    \mathrm{SE}(3) := \left\{ \mathbf{T} = \left[ \begin{array}{cc}
        \mathbf{R}~ & \mathbf{t} \\
        \mathbf{0}^\top & 1
    \end{array} \right] \in \mathbb{R}^{4 \times 4} \middle| \mathbf{R} \in \mathrm{SO}(3), \mathbf{t} \in \mathbb{R}^3\right\},
\end{equation}
where $\mathrm{SO}(3)$ is the \textit{Special Orthogonal group} is the space of valid rotation matrices:
\begin{equation}
    \mathrm{SO}(3) := \left\{ \mathbf{R} \in \mathbb{R}^{3 \times 3} | \mathbf{R}\mathbf{R}^\top = \mathbf{I}_3, \operatorname{det}\mathbf{R} = 1 \right\},~~~
\end{equation}
and $\mathbf{I}_d$ is the identity matrix of dimension $d$. 
A variety of methods have been developed for parameterizing these objects such as Euler angles or quaternions  \cite{spong2005robot}.  

\begin{figure}[t]
    \centering
    \includegraphics[width=0.7\columnwidth]{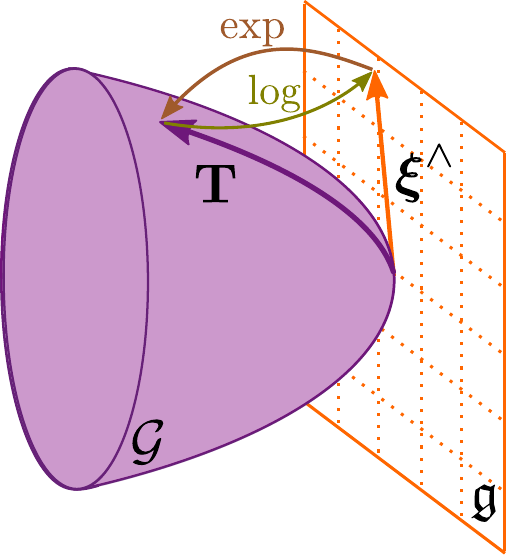}
    \caption{
    The Lie algebra, $\mathfrak{g}$, is the tangent space to the Lie group, $\mathcal{G}$, centered at the identity element.
    The Lie algebra represents the space of all possible velocities a particle at a given point on the group could take. 
    The \textit{exponential map} maps velocities in the Lie algebra, $\boldsymbol{\xi}^\wedge \in \mathfrak{g}$, to their associated action in the Lie group, $\mathbf{T} \in \mathcal{G}$, and the \textit{logarithm map} performs the inverse operation.}
    \label{fig:lie_algebra}
\end{figure}

Both $\mathrm{SE}(3)$ and $\mathrm{SO}(3)$ are matrix \textit{Lie groups} meaning they are smooth manifolds that also satisfy the formal definition of a mathematical group~\cite{tapp2016matrix,baker2012matrix} with the standard matrix multiplication operation. 
Intuitively, this means that while the general group is non-linearm, each of these groups can be locally approximated using a Euclidean vector space. 
Additionally, for any given point on the manifold, consider the set of all paths on the manifold that pass through that point. 
The set of all velocities (both in terms of direction and speed) of those paths at the given point form a vector space called the tangent space. 
The tangent space centered at the identity is called the \textit{Lie algebra}. 
This relationship is depicted in \figref{fig:lie_algebra}.

The Lie algebra of $\mathrm{SO}(3)$ is denoted $\mathfrak{so}(3)$ and is the space of skew-symmetric $3\times3$ matrices~\cite{chirikjian2011stochastic,tapp2016matrix}:
\begin{equation}
    \mathfrak{so}(3) := \left\{ \boldsymbol{\omega} \in \mathbb{R}^{3 \times 3} ~| ~\boldsymbol{\omega}^\top = - \boldsymbol{\omega}\}\right.~~~
\end{equation}
This space is isomorphic to $\mathbb{R}^3$ since skew-symmetric matrices have zeros on the diagonal entries and the lower entries are completely identified by three upper entries (hence $\mathrm{dim}~\mathfrak{so}(3) = 3$). Therefore, it is very convenient to work with $\mathbb{R}^3$ instead. 
Note, we can use the $\wedge$ operator to take an element of $\mathbb{R}^3$ and transform it to an element of $\mathfrak{so}(3)$:
\begin{equation}
    \boldsymbol{\phi}^\wedge := \left[ \begin{array}{c}
         \phi_{1} \\
         \phi_{2} \\
         \phi_{3}
    \end{array} \right]^\wedge = 
    \left[ \begin{array}{ccc}
        0 & -\phi_{3} & \phi_{2} \\
        \phi_{3} & 0 & -\phi_{1} \\
        -\phi_{2} & \phi_{1} & 0
    \end{array} \right] \in \mathfrak{so}(3),
\end{equation}
where $\boldsymbol{\phi} \in \mathbb{R}^3$.
The $\vee$ operator denotes the inverse of $\wedge$.

Similarly, the Lie algebra of $\mathrm{SE}(3)$, or $\mathfrak{se}(3)$, is defined as follows:
\begin{equation}
    \mathfrak{se}(3) := 
    \left\{ \left[ \begin{array}{cc}
        \boldsymbol{\omega} & \boldsymbol{\rho}  \\
        \mathbf{0}^\top & 0
    \end{array} \right]~ \middle| ~\boldsymbol{\omega} \in \mathfrak{so}(3), \boldsymbol{\rho} \in \mathbb{R}^3 \right\}.
\end{equation}
This space of matrices is isomorphic to $\mathbb{R}^6$ and we overload the $\wedge$ operator
to convert between the Euclidean vector and matrix forms:
\begin{equation}
    \boldsymbol{\xi}^\wedge := \left[ \begin{array}{c}
         \boldsymbol{\rho} \\
          \boldsymbol{\phi}
    \end{array} \right]^\wedge = 
    \left[ \begin{array}{cc}
        \boldsymbol{\phi}^\wedge & \boldsymbol{\rho}  \\
        \mathbf{0}^\top & 0
    \end{array} \right] \in \mathfrak{se}(3),
\end{equation}
where $\boldsymbol{\xi} \in \mathbb{R}^6$ and $\boldsymbol{\rho}, \boldsymbol{\phi} \in \mathbb{R}^3$.

Understanding the relationship between the group and algebra spaces can enable one to leverage the fact that the algebra is a vector space.
The next few subsections cover some important concepts from Lie group Theory that we need in the rest of the paper. 
In doing so, we use $\mathcal{G}$ to represent a given Lie group and  $\mathfrak{g}$ to represent its associated Lie algebra.

\subsection{The Exponential Map}

The Lie algebra, $\mathfrak{g}$, represents the tangent space of the manifold at the identity. 
However, given a specific tangent vector, we may want to 
convert it to its associated transformation in the group space $\mathcal{G}$.
The \textit{exponential map}, $\operatorname{exp} : \mathfrak{g} \rightarrow \mathcal{G}$, (which can be defined in closed form for $\mathrm{SE}(3)$), enables us to perform this conversion,
\begin{equation}
    \operatorname{exp}(\boldsymbol{\xi}^\wedge) = \sum^\infty_{k=0} \frac{ (\boldsymbol{\xi}^\wedge)^k }{k!} = \mathbf{I}_4 + \boldsymbol{\xi}^\wedge + \frac{(\boldsymbol{\xi}^\wedge)^2}{2} + \cdots.
\end{equation}
The \textit{logarithm map}, $\operatorname{log} : \mathcal{G} \rightarrow \mathfrak{g}$, on the other hand, enables us to go the other direction from an action/transformation in the group space to the velocity that would have induced it,
\begin{equation}
    \operatorname{log}(\mathbf{T}) = \sum^\infty_{k=1} (-1)^{k+1} \frac{(\mathbf{T} - \mathbf{I}_4)^k}{k}.
\end{equation}
This relationship is visualized in \figref{fig:lie_algebra}.

\subsection{The Adjoint Action}

Assuming $\mathbf{T} \in \mathcal{G}$ and $\boldsymbol{\xi} \in \mathfrak{g}$, the adjoint action of $\mathbf{T}$ on $\boldsymbol{\xi}$, or $\operatorname{Ad}_\mathbf{T}(\boldsymbol{\xi})$ is defined as follows:
\begin{equation}
    \operatorname{Ad}_\mathbf{T}(\boldsymbol{\xi}) := \operatorname{Ad}_\mathbf{T} \boldsymbol{\xi} = \operatorname{log}( \mathbf{T} \operatorname{exp} ( \boldsymbol{\xi} ) \mathbf{T}^{-1} ).
\end{equation}
The adjoint action describes the affect that transforming to the 
group space applying a transformation on the left and its inverse 
on the right has on an element of the Lie algebra. 
This gives rise to the following property, which we use later on:
\begin{align}
    \mathbf{T} \operatorname{exp} ( \boldsymbol{\xi} ) = \operatorname{exp} ( \operatorname{Ad}_\mathbf{T} \boldsymbol{\xi} ) \mathbf{T} \label{eq:adjoint_property}
\end{align}


\subsection{The Baker-Campbell-Hausdorff (BCH) Formula}

Finally, we also want to characterize the effect that multiplication in the group space has on the Lie algebra.
More specifically, suppose we want to compute the Lie algebra element, $\boldsymbol{\xi}_{ac}\in \mathfrak{g}$, that is generated by taking the logarithm of the product of the exponential of two Lie algebra elements $\boldsymbol{\xi}_{ab}, \boldsymbol{\xi}_{bc} \in \mathfrak{g}$.
The \textit{Baker-Campbell-Hausdorff (BCH)} formula describes this relationship purely in the Lie algebra space without requiring the application of the exponential or the logarithm:
\begin{align}
    \boldsymbol{\xi}_{ac} &= 
    \operatorname{log}( \operatorname{exp}(\boldsymbol{\xi}_{ab}) \operatorname{exp}(\boldsymbol{\xi}_{bc})) \label{eq:BCH}\\
    &= \boldsymbol{\xi}_{ab} + \boldsymbol{\xi}_{bc} + \frac{1}{2}[\boldsymbol{\xi}_{ab}, \boldsymbol{\xi}_{bc}] + \nonumber \\ 
    & + \frac{1}{12}([\boldsymbol{\xi}_{ab}, [\boldsymbol{\xi}_{ab}, \boldsymbol{\xi}_{bc}]] + [\boldsymbol{\xi}_{bc},[\boldsymbol{\xi}_{bc},\boldsymbol{\xi}_{ab}]]) + \cdots\nonumber,
\end{align}
where $[\cdot, \cdot]$ is the Lie bracket of $\mathfrak{g}$ \cite[(7.18)]{barfoot2017state}. 
For brevity, in the special case where we consider elements in $\mathrm{SE}(3)$, we adopt the notation of \cite{barfoot2014associating}:
\begin{equation}
    \boldsymbol{\xi}_i^\curlywedge := \left[ \begin{array}{c}
         \boldsymbol{\rho}_i \\
          \boldsymbol{\phi}_i
    \end{array} \right]^\curlywedge = 
    \left[ \begin{array}{cc}
        \boldsymbol{\phi}_i^\wedge & \boldsymbol{\rho}_i^\wedge  \\
        \mathbf{0} &  \boldsymbol{\phi}_i^\wedge
    \end{array} \right],
\end{equation}
which allows us to write the BCH formula as:
\begin{align}
    \xib{ac} &= \xib{ab} + \xib{bc} + \frac{1}{2}\xic{ab}\xib{bc} + \frac{1}{12}\xic{ab}\xic{ab}\xib{bc} + \frac{1}{12}\xic{bc}\xic{bc}\xib{ab} \nonumber\\
    &- \frac{1}{24}\xic{bc}\xic{ab}\xic{ab}\xib{bc} + \dots. \label{eq:BCH_SE3}
\end{align}

\subsection{Defining Random Variables over Poses}

Finally, as discussed in \cite{barfoot2014associating}, one can 
define random variables for $\mathrm{SE}(3)$ according to 
\begin{equation}
    \mathbf{T}_\ell := \operatorname{exp}(\boldsymbol{\xi}_\ell^{\wedge}) \bar{\mathbf{T}}_\ell
    \label{eq:T_i}
\end{equation}
where $\bar{\mathbf{T}}_\ell\in \mathrm{SE}(3)$ is a `large' noise free value and $\boldsymbol{\xi}_\ell \in \mathbb{R}^6$ is a `small' noisy perturbation (using the nomenclature
of \cite{barfoot2014associating}).
Two examples of this noisy perturbation are depicted in the middle row of \figref{fig:main_example}.
By defining $\boldsymbol{\xi}_\ell$ to be a zero-mean Gaussian random variable $\boldsymbol{\xi}_\ell \sim \mathcal{N}(\mathbf{0}, \mathbf{\Sigma}_\ell)$ in the Lie algebra, we induce a probability distribution function over $\mathrm{SE}(3)$ that is parameterized with a mean $\bar{\mathbf{T}}_\ell \in \mathrm{SE}(3)$ and a covariance $\mathbf{\Sigma}_\ell$ defined in the Lie algebra \cite{barfoot2014associating}. 

\begin{figure*}[t]
    \centering
    \includegraphics[width=\textwidth]{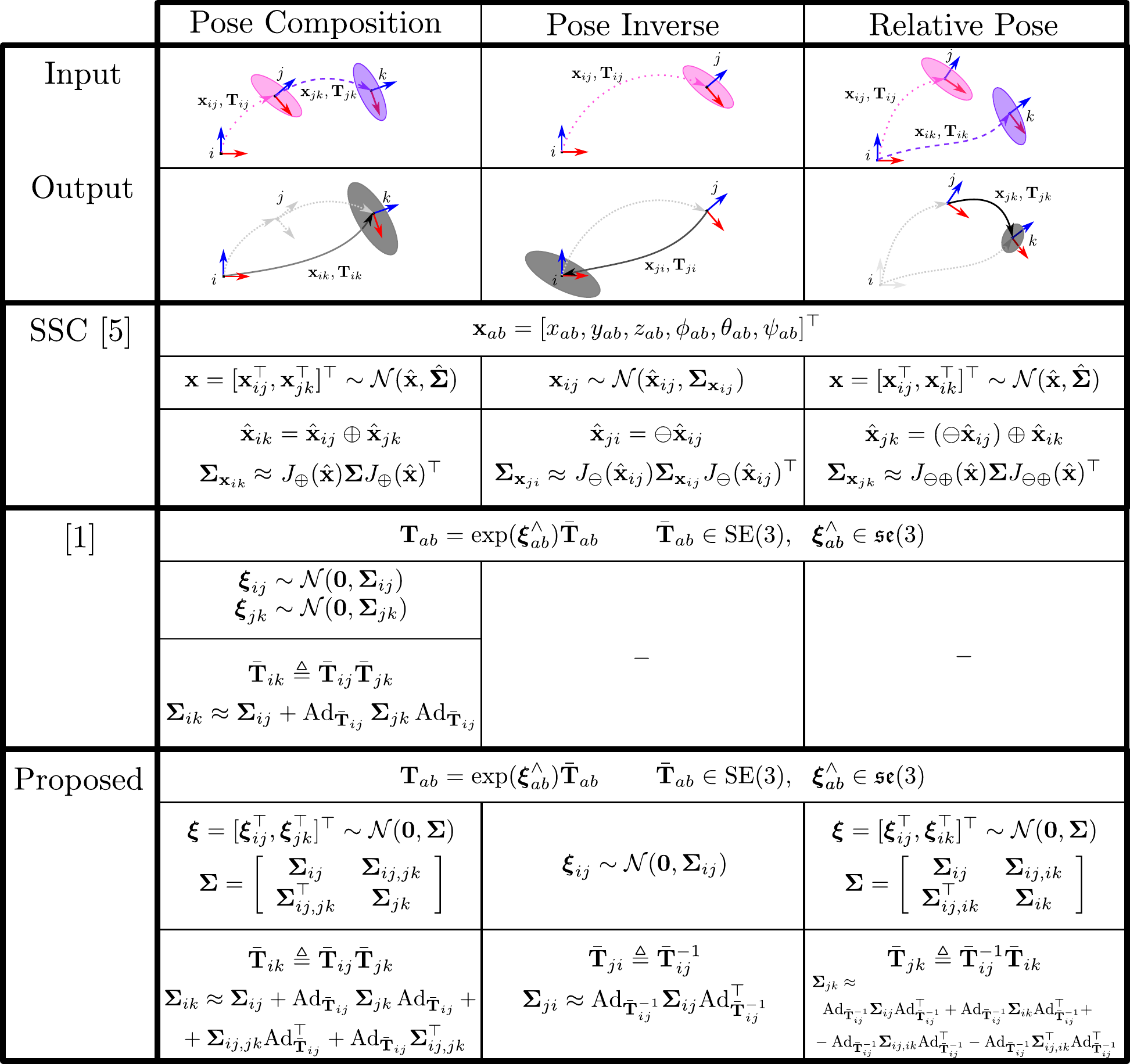}
    \caption{
    Summary of the pose composition, pose inverse, and relative pose operations and their corresponding uncertainty propagation methods as proposed by \citet{smith1990a}, \citet{barfoot2014associating}, and in this paper. The indices $i$, $j$, and $k$ correspond to specific coordinate frames of the robotic vehicle at different time steps or locations. 
    SSC \cite{smith1990a} parameterizes the pose of frame $b$ with respect to frame $a$ using a vector of Euler angle and translation parameters $\mathbf{x}_{ab}$. They then model this vector of parameters as being drawn from a multivariate Guassian distribution with mean $\hat{\mathbf{x}}$ and covariance $\boldsymbol{\Sigma}$. Under this model, they derive first order uncertainty propagation formulas for the pose composition, pose inverse, and relative pose operations. However, this coordinate based parameterization is unable to accurately model pose uncertainty because the parameter vector $\mathbf{x}_{ab}$ is not truly Gaussian. 
    \citet{barfoot2014associating} instead parameterize the pose of frame $b$ with respect to frame $a$, using a mean element of the Special Euclidean group, $\bar{\mathbf{T}}_{ab}$, and an uncertain perturbation or noise parameter $\boldsymbol{\xi}_{ab}^\wedge$ defined 
    in the Lie algebra $\mathfrak{se}(3)$. This enables them to model $\boldsymbol{\xi}_{ab}$ using a Gaussian distribution and accurately take into account the non-linear structure of the group. However, \cite{barfoot2014associating} assumes the poses are independent and focuses primarily on the pose composition operation. 
    The \textbf{primary contribution} of this paper is the extension of the method presented by  \citet{barfoot2014associating} to jointly correlated poses as well as the derivation of 
    the pose inverse and relative pose operations while taking advantage of the Lie algebra to characterize uncertainty.} 
    \label{fig:summary}
\end{figure*}

We use these properties in Sections \ref{sec:lie_joint_uncertainty}, \ref{sec:pose_composition}, \ref{sec:inverse}, and \ref{sec:relative_pose} to propose a 
framework for modeling jointly correlated poses and to derive
uncertainty propagation formulas for the operations in \figref{fig:summary}. However, we first review the SSC coordinate based method.

\section{Review of SSC}
\label{sec:SSC}

The stochastic map, proposed by \citet{smith1990a}, consists of multiple uncertain spatial relationships that are treated as jointly Gaussian multivariate random variables and parameterized using a mean vector of positions and Euler angles and an associated covariance matrix. \citet{smith1990a}  also proposes three operations that can extract information from this map that may not have been directly estimated.
These three operations are pose composition (head-to-tail in \cite{smith1990a}), pose inverse, and relative pose estimation (tail-to-tail in \cite{smith1990a}), as shown in \figref{fig:summary}. This section reviews the pose representation and SSC operations formulated in \cite{smith1990a}.

\FloatBarrier

\subsection{Pose Representation and the \textit{Stochastic Map}}

Under SSC notation, the relative transformation between the coordinate frame $i$ and $j$, or the pose of coordinate frame $j$ with respect to frame $i$, is denoted by
\begin{equation}
    \mathbf{x}_{ij} = [x_{ij}, y_{ij}, z_{ij}, \phi_{ij}, \theta_{ij}, \psi_{ij}]^\top, \label{eq:SSC_rep}
\end{equation}
where $x_{ij}$, $y_{ij}$, and $z_{ij}$ are position coordinates and $\phi_{ij}$, $\theta_{ij}$, and $\psi_{ij}$ are Euler angles that encode the position and orientation of the coordinate frame $j$ with respect to the frame $i$.
However, because each relative transformation is derived from noisy measurements, our estimate is uncertain and we do not know the true value of $\mathbf{x}_{ij}$.
Instead, we track a nominal mean value of its parameters and an associated $6\times 6$ covariance matrix, $\hat{\mathbf{x}}_{ij}$ and $\mathbf{\Sigma}_{\mathbf{x}_{ij}}$, respectively.

The \textit{stochastic map}, proposed in \cite{smith1990a}, treats all of the uncertain transformations as jointly Gaussian multivariate random variables by stacking them into a single state vector $\mathbf{x}$ and tracking the mean and covariance of that state. 
Assuming we have $n$ relative transformations we want to track, if we index them from $a=1, \dots, n$, then $\mathbf{x}$ and its associated mean and covariance are:
\begin{align}
\mathbf{x} &= 
    \left[ \begin{array}{c} 
        \mathbf{x}_1 \\ \vdots \\ \mathbf{x}_n \end{array} \right],
\hat{\mathbf{x}} =       
    \left[ \begin{array}{c} 
        \hat{\mathbf{x}}_1 \\ \vdots \\ \hat{\mathbf{x}}_n \end{array} \right], 
\hat{\mathbf{\Sigma}} =       
    \left[ \begin{array}{ccc} 
        \mathbf{\Sigma}_{\mathbf{x}_1} & \dots & \mathbf{\Sigma}_{\mathbf{x}_1 \mathbf{x}_n} \\
        \vdots & \ddots & \vdots \\
        \mathbf{\Sigma}_{\mathbf{x}_n \mathbf{x}_1} & \dots  & \mathbf{\Sigma}_{\mathbf{x}_n} \\
        \end{array}\right] \label{eq:SSC_joint_rep}
\end{align}
where $\mathbf{x}$ is a vector of length $6n$, $\mathbf{\Sigma}_{\mathbf{x}_a}$ is the $6\times6$ covariance matrix of the relative transformation $\mathbf{x}_a$ and the off diagonal blocks of the form $\mathbf{\Sigma}_{\mathbf{x}_a \mathbf{x}_b}$ are the respective $6\times6$ cross covariance matrices.

\subsection{Pose Composition (Head-to-Tail)}
\label{sec:SSC:head-to-tail}

Suppose we are given a noisy observation of a robot's relative pose between time steps $i$ and $j$ $(\mathbf{x}_{ij})$ and another observation of its relative pose between time steps $j$ and $k$ $(\mathbf{x}_{jk})$, we may want to calculate the relative pose between time steps $i$ and $k$ $(\mathbf{x}_{ik})$ by composing the observations $\mathbf{x}_{ij}$ and $\mathbf{x}_{jk}$ (see \figref{fig:summary}).
The SSC \textit{head-to-tail} operation is a nonlinear function $f_\oplus: \mathbb{R}^{6} \times \mathbb{R}^6 \rightarrow \mathbb{R}^6$ that takes the parameter vectors $\mathbf{x}_{ij}$ and $\mathbf{x}_{jk}$ as input and outputs the parameter vector $\mathbf{x}_{ik}$ that results from composing the respective homogeneous transformation matrices. 
The $\oplus$ operator denotes this operation:
\begin{equation}
    \mathbf{x}_{ik} \triangleq \mathbf{x}_{ij} \oplus \mathbf{x}_{jk} = f_\oplus(\mathbf{x}_{ij}, \mathbf{x}_{jk}).
\end{equation}
The mean and covariance of the resulting pose are estimated up-to first order as
\begin{equation}
    \hat{\mathbf{x}}_{ik} = \hat{\mathbf{x}}_{ij} \oplus \hat{\mathbf{x}}_{jk}
\end{equation}
and 
\begin{equation}
    \mathbf{\Sigma}_{\mathbf{x}_{ik}} \approx J_\oplus(\hat{\mathbf{x}}_{ij}, \hat{\mathbf{x}}_{jk}) ~ \hat{\mathbf{\Sigma}} ~
    J_\oplus(\hat{\mathbf{x}}_{ij}, \hat{\mathbf{x}}_{jk})^\top
\end{equation}
where $J_\oplus(\hat{\mathbf{x}}_{ij}, \hat{\mathbf{x}}_{jk})$ is the Jacobian of $f_\oplus$ at $\hat{\mathbf{x}}_{ij}$ and $\hat{\mathbf{x}}_{jk}$, and 
\begin{align}
    \hat{\mathbf{\Sigma}} 
    &= 
    \left[ \begin{array}{cc}
        \mathbf{\Sigma}_{
        \mathbf{x}_{ij}} & ~~~\mathbf{\Sigma}_{\mathbf{x}_{ij}\mathbf{x}_{jk}}  \\
        ~~~\mathbf{\Sigma}_{\mathbf{x}_{ij}\mathbf{x}_{jk}}^\top & \mathbf{\Sigma}_{\mathbf{x}_{jk}}
    \end{array} \right]. 
\end{align} 

\subsection{Pose Inverse}
\label{sec:SSC:inverse}

Suppose we are given a robotic vehicle that is tasked with navigating through an \textit{a priori} unknown environment after which it must return to the origin location. 
While it is common to represent the robotic vehicle's current location with respect to the origin using estimation theory, it may be useful to instead characterize the pose of the origin with respect to the local robot coordinate frame. 
Formally, given an uncertain estimate of the pose of coordinate frame $j$ with respect to frame $i$ ($\mathbf{x}_{ij}$), we want to determine the pose of frame $i$ with respect to frame $j$ ($\mathbf{x}_{ji}$). 
This amounts to finding the vector $\hat{\mathbf{x}}_{ji}$ that corresponds to the inverse of $\hat{\mathbf{x}}_{ij}$ (in terms of homogeneous transformation matrices) and representing the uncertainty with respect to this new frame of reference. 

The SSC \textit{pose inverse} operation is a nonlinear function $f_\ominus: \mathbb{R}^6 \rightarrow \mathbb{R}^6 $ takes a pose and computes the inverse of the pose as a homogeneous transformation matrix: 
\begin{equation}
    \mathbf{x}_{ji} \triangleq \ominus \mathbf{x}_{ij} = f_\ominus(\mathbf{x}_{ij}).
\end{equation}
The mean and covariance of the resulting pose are estimated up-to first order as
\begin{equation}
    \hat{\mathbf{x}}_{ji} = \ominus \hat{\mathbf{x}}_{ij} 
\end{equation}
and 
\begin{equation}
    \mathbf{\Sigma}_{\mathbf{x}_{ji}} \approx J_\ominus(\hat{\mathbf{x}}_{ij}) ~ \mathbf{\Sigma}_{\mathbf{x}_{ij}} ~
    J_\ominus(\hat{\mathbf{x}}_{ij})^\top
\end{equation}
where $J_\ominus(\hat{\mathbf{x}}_{ij})$ is the Jacobian of $f_\ominus$ at $\hat{\mathbf{x}}_{ij}$.

\subsection{Relative Pose (Tail-to-Tail)}
\label{sec:SSC:tail-to-tail}

Finally, the SSC \textit{tail-to-tail} operation takes uncertain estimates of two coordinate frames with respect to a single origin frame and evaluates the relative pose between them. 
More succinctly, given $\mathbf{x}_{ij}$ and $\mathbf{x}_{ik}$, find $\mathbf{x}_{jk}$. 
This operation is useful when using a method such as pose graph \ac{SLAM} for navigation since the robot pose at multiple time steps is often estimated with respect to a single fixed coordinate frame. 

The SSC \textit{relative pose} operation is a nonlinear function $f_{\ominus\oplus}: \mathbb{R}^6 \times \mathbb{R}^6 \rightarrow \mathbb{R}^6$ that is defined by first applying the inverse and then the head-to-tail operation:
\begin{equation}
    \mathbf{x}_{jk} \triangleq (\ominus \mathbf{x}_{ij}) \oplus \mathbf{x}_{ik} = f_{\ominus\oplus}(\mathbf{x}_{ij}, \mathbf{x}_{ik}).
\end{equation}
The mean and covariance of the resulting pose up-to first order are estimated as
\begin{equation}
    \hat{\mathbf{x}}_{jk} = (\ominus \hat{\mathbf{x}}_{ij}) \oplus \hat{\mathbf{x}}_{ik}
\end{equation}
and 
\begin{equation}
    \mathbf{\Sigma}_{\mathbf{x}_{jk}} \approx J_{\ominus\oplus}(\hat{\mathbf{x}}_{ij}, \hat{\mathbf{x}}_{ik}) ~ \hat{\mathbf{\Sigma}} ~
    J_{\ominus\oplus}(\hat{\mathbf{x}}_{ij}, \hat{\mathbf{x}}_{ik})^\top
\end{equation}
where 
$J_{\ominus\oplus}(\hat{\mathbf{x}}_{ij}, \hat{\mathbf{x}}_{ik})$ is the Jacobian of $f_{\ominus\oplus}$ at $\hat{\mathbf{x}}_{ij}$ and $\hat{\mathbf{x}}_{ik}$, and 
\begin{align}
    \hat{\mathbf{\Sigma}} 
    &= 
    \left[ \begin{array}{cc}
        \mathbf{\Sigma}_{
        \mathbf{x}_{ij}} & ~~~\mathbf{\Sigma}_{\mathbf{x}_{ij}\mathbf{x}_{ik}}  \\
        ~~~\mathbf{\Sigma}_{\mathbf{x}_{ij}\mathbf{x}_{ik}}^\top & \mathbf{\Sigma}_{\mathbf{x}_{ik}}
    \end{array} \right]. 
\end{align}

\section{Jointly Characterizing Uncertainty in the Lie algebra}
\label{sec:lie_joint_uncertainty}


While the SSC operations proposed in \cite{smith1990a} have been used widely since they were introduced, over the past decade Lie algebra based methods have been shown to provide a more accurate characterization of uncertainty \cite{barfoot2014associating, kim2017uncertainty}. 
However, while recent years have seen an increase in use of Lie algebra based  methods for uncertainty propagation \cite{forster2017manifold, rhartley-2018a, wheeler2018relative}, existing methods assume that individual measurements are independent \cite{barfoot2014associating}, which may not be the case when the underlying pose estimates are derived from a \ac{SLAM} solution. 
We now describe how this assumption can be dropped and poses can be modeled as jointly correlated within the Lie algebra space.

Assume that we have $n$ uncertain poses  $\{\mathbf{T}_1, \dots, \mathbf{T}_n\}$, each of which is defined according to \eqref{eq:T_i}. 
If we assume that the poses are statistically independent, then we can parameterize this distribution of poses with the set of associated mean and covariance matrices $\{\bar{\mathbf{T}}_1, \mathbf{\Sigma}_1, \dots, \bar{\mathbf{T}}_n, \mathbf{\Sigma}_n\}$. 
However, if the uncertainty associated with the set of poses is correlated, then this can be modeled by concatenating the set of perturbation vectors $\{ \boldsymbol{\xi}_{1}, \dots, \boldsymbol{\xi}_n \} $ into a single vector $\boldsymbol{\xi}_{1:n} \in \mathbb{R}^{6n}$ and represent the uncertainty of the distribution using a single covariance matrix:
\begin{equation}
    \boldsymbol{\xi}_{1:n} = \left[ \begin{array}{c}
        \boldsymbol{\xi}_{1}  \\
        \vdots \\
        \boldsymbol{\xi}_n
    \end{array} \right], ~~~
    \mathbf{\Sigma}_{1:n} = \left[ \begin{array}{ccc}
        \mathbf{\Sigma}_1  & \dots, & \mathbf{\Sigma}_{1,n}  \\
        \vdots & \ddots & \vdots \\
        \mathbf{\Sigma}_{1,n}^\top & \dots & \mathbf{\Sigma}_{n}
    \end{array} \right], \label{eq:lie_joint_rep_cov}
\end{equation}
where $\boldsymbol{\xi}_{1:n} \sim \mathcal{N}(\mathbf{0}, \mathbf{\Sigma}_{1:n})$. 
Thus, we can parameterize the distribution of $n$ poses using the set of each of their means 
\begin{equation}
    \bar{\mathbf{T}}_{1:n} = \{ \bar{\mathbf{T}}_1, \dots, \bar{\mathbf{T}}_n\} \label{eq:lie_joint_rep_means}
\end{equation} 
and the covariance matrix $\mathbf{\Sigma}_{1:n}$. 
Note that this only extends since we define the uncertainty in the Lie algebra. 
The next three sections describe how to apply the composition, inverse, and relative pose operations on uncertain poses represented in this manner.

\section{Jointly Distributed Pose Composition}
\label{sec:pose_composition}

This section describes how to compose two uncertain poses who's associated perturbation vectors are jointly Gaussian in the Lie algebra as described in the last section. 
This is the Lie group-based or coordinate free equivalent to the SSC head-to-tail operation described in Section \ref{sec:SSC:head-to-tail}. 

\subsection{Pose Composition Operation Derivation}

Suppose we have two uncertain poses $\mathbf{T}_{ij}$ and $\mathbf{T}_{jk}$ with perturbations $\boldsymbol{\xi}_{ij}$ and $\boldsymbol{\xi}_{jk}$ that are jointly Gaussian in the Lie algebra with covariance matrix
\begin{align}
    \mathbf{\Sigma} 
    &= 
    \left[ \begin{array}{cc}
        \mathbf{\Sigma}_{ij} & ~~~\mathbf{\Sigma}_{ij,jk}  \\
        ~~~\mathbf{\Sigma}_{ij,jk}^\top & \mathbf{\Sigma}_{jk}
    \end{array} \right]. \nonumber
\end{align}
Our goal is to find the mean and covariance of $\mathbf{T}_{ik}$, $\{\bar{\mathbf{T}}_{ik}, \mathbf{\Sigma}_{ik}\}$.
Under the standard group multiplication operation:
\begin{equation}
    \mathbf{T}_{ik} = \mathbf{T}_{ij} \mathbf{T}_{jk}.
\end{equation}
Following the random variable definition in \eqref{eq:T_i} and using the property described in \eqref{eq:adjoint_property},
\begin{align}
    \operatorname{exp}(\boldsymbol{\xi}_{ik}^\wedge) \bar{\mathbf{T}}_{ik} 
    &= 
    \operatorname{exp}(\boldsymbol{\xi}_{ij}^\wedge) \bar{\mathbf{T}}_{ij}
    \operatorname{exp}(\boldsymbol{\xi}_{jk}^\wedge) \bar{\mathbf{T}}_{jk} \nonumber\\
    &= 
    \operatorname{exp}(\boldsymbol{\xi}_{ij}^\wedge) \operatorname{exp}((\mathrm{Ad}_{\bar{\mathbf{T}}_{ij}}\boldsymbol{\xi}_{jk})^\wedge)
    \bar{\mathbf{T}}_{ij} \bar{\mathbf{T}}_{jk}
\end{align}
where $\mathrm{Ad}_{\bar{\mathbf{T}}_{ij}}$ is the matrix form of the adjoint action of $\bar{\mathbf{T}}_{ij}$ on $\mathfrak{se}(3)$.
Letting
\begin{equation}
    \bar{\mathbf{T}}_{ik} \triangleq \bar{\mathbf{T}}_{ij} \bar{\mathbf{T}}_{jk}, \label{eq:pose_compound_mu}
\end{equation}
gives us
\begin{align}
    \operatorname{exp}(\boldsymbol{\xi}_{ik}^\wedge) 
    &= \operatorname{exp}(\boldsymbol{\xi}_{ij}^\wedge) \operatorname{exp}((\mathrm{Ad}_{\bar{\mathbf{T}}_{ij}}\boldsymbol{\xi}_{jk})^\wedge).
\end{align}

We can now use the \ac{BCH} formula \eqref{eq:BCH_SE3} to show that
\begin{align}
    \xib{ik} &= \xib{ij} + \xip{jk} + \frac{1}{2}\xic{ij}\xip{jk} + \frac{1}{12}\xic{ij}\xic{ij}\xip{jk} + \frac{1}{12}\xipc{jk}\xipc{jk}\xib{ij} + \nonumber\\
    &- \frac{1}{24}\xipc{jk}\xic{ij}\xic{ij}\xip{jk} + \dots \label{eq:BCH_compound_pose}
\end{align}
where $\xip{jk} = \mathrm{Ad}_{\bar{\mathbf{T}}_{ij}}\boldsymbol{\xi}_{jk}$.
Computing the covariance $\mathbf{\Sigma}$ amounts to evaluating $E[\xib{ik}\xib{ik}^\top]$.
Multiplying out up-to fourth order, we get: 
\begin{align}
    E[\xib{ik}\xibt{ik}] &\approx E\left[ \underbrace{\xib{ij}\xibt{ij} + 
    \xip{jk}\xipt{jk}}_\text{2nd Order Diag. Terms} + \underbrace{\xib{ij}\xipt{jk} + \xip{jk}\xibt{ij}}_\text{2nd Order Cross Terms} +  \right. \label{eq:2nd_order}\\
    + &\frac{1}{12}( 
    (\xic{ij}\xic{ij})(\xip{jk}\xipt{jk}) + 
    (\xip{jk}\xipt{jk})(\xic{ij}\xic{ij})^\top + \nonumber \\
    + &~~~~~(\xipc{jk}\xipc{jk})(\xib{ij}\xibt{ij}) + 
    (\xib{ij}\xibt{ij})(\xipc{jk}\xipc{jk})^\top ) + \label{eq:4th_order_diag}\\
    + &\underbrace{\frac{1}{4}\xic{ij}(\xip{jk}\xipt{jk})\xict{ij} + ~~~~~~~~~~~~~~~~~~~~~~~~~~~~~}_\text{4th Order Diagonal Terms} \nonumber \\
    + &\frac{1}{12}( 
    (\xib{ij}\xipt{jk})(\xict{ij}\xict{ij}) + 
    (\xip{jk}\xibt{ij})(\xipct{jk}\xipct{jk}) + \label{eq:4th_order_cross} \\ 
    + &\left.\underbrace{~~~~~(\xic{ij}\xic{ij})(\xip{jk}\xibt{ij}) +
    (\xipc{jk}\xipc{jk})(\xib{ij}\xipt{jk}))~~~~~}_\text{4th Order Cross Terms} 
    \right]. \nonumber 
\end{align}

This derivation is identical to the one proposed in \cite{barfoot2014associating} until 
the last step. 
Since \cite{barfoot2014associating} assumes the individual poses are independent, the cross terms in \eqref{eq:2nd_order} and \eqref{eq:4th_order_cross} are zero and the evaluation of the 4th order terms in \eqref{eq:4th_order_diag} is simplified. 
If this assumption is true and $\xib{ij}$ and $\xip{jk}$ are independent of one another, as is the case when $\mathbf{T}_{ij}$ and $\mathbf{T}_{jk}$ are independent measurements of consecutive robot motion, then the cross terms in \eqref{eq:4th_order_diag} and the covariance $\mathbf{\Sigma}$ can be calculated up to 4th order as described in \cite{barfoot2014associating}.
However, if the poses $\mathbf{T}_{ij}$ and $\mathbf{T}_{jk}$ are correlated, as is often the case when they are derived from the solution of a \ac{MLE} problem such as Pose Graph SLAM or in the case of wheel slip, the cross terms must be  included and evaluation of the fourth order terms in \eqref{eq:4th_order_diag} and \eqref{eq:4th_order_cross} becomes more difficult\footnote{If increased accuracy is needed then Isserlis Theorem can be applied to evaluate the fourth order terms in \eqref{eq:4th_order_diag} and \eqref{eq:4th_order_cross}.}. 
If the correlation is not taken into account, then the result of the pose composition operation under-approximates the true distribution and consistency is lost as shown in \figref{fig:compose_example}.

A first order estimate of covariance (second order in the perturbation variables) can
be obtained be evaluating
\begin{align}
    E[\xib{ik}\xibt{ik}] \approx &E [ \xib{ij}\xibt{ij} ] + E[\xip{jk}\xipt{jk}] ~+  \\       
    &+ E[\xib{ij}\xipt{jk}] + E[\xip{jk}\xibt{ij}], \nonumber
\end{align} 
resulting in
\begin{align}
    \mathbf{\Sigma}_{ik} \approx &\mathbf{\Sigma}_{ij} + \mathrm{Ad}_{\bar{\mathbf{T}}_{ij}} \mathbf{\Sigma}_{jk} \mathrm{Ad}_{\bar{\mathbf{T}}_{ij}}^\top + \label{eq:pose_compound_cov}\\ &+\mathbf{\Sigma}_{ij,jk} \mathrm{Ad}_{\bar{\mathbf{T}}_{ij}}^\top +\mathrm{Ad}_{\bar{\mathbf{T}}_{ij}} \mathbf{\Sigma}_{ij,jk}^\top . \nonumber
\end{align}
Thus, first order pose composition on the Lie algebra can be performed for correlated poses
using \eqref{eq:pose_compound_mu} for mean propagation and \eqref{eq:pose_compound_cov} for covariance propagation.
As noted previously, this is equivalent to the first order method presented
in \cite{barfoot2014associating} with the exception of the two additional 
cross terms in \eqref{eq:pose_compound_cov} needed to model cross correlation.


\subsection{Accurately Modeling Group Structure}
\label{sec:pose_composition:comparison}

\begin{figure}[t]
    \centering
    \includegraphics[width=\columnwidth]{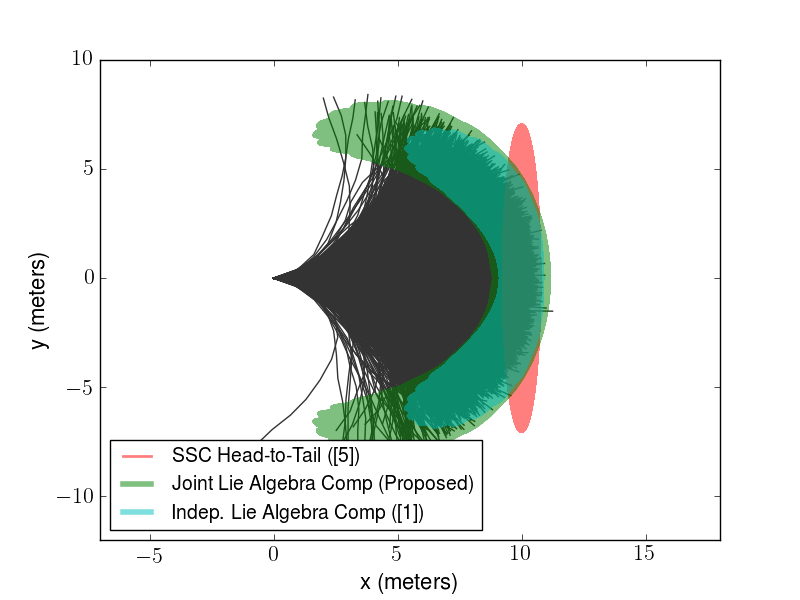}
    \caption{
    Plots of 10000 sample trajectories each made up of 10 noisy pose transformations. The 95\% likely uncertainty ellipse predicted by first order uncertainty propagation through the SSC head-to-tail operation is shown in red, while a representation of the flattened 95\% likely uncertainty position ellipsoids predicted by the Lie algebra pose composition methods when correlation is and is not taken into account are shown in green and cyan, respectively. }
    \label{fig:compose_example}
\end{figure}

While both the SSC operations presented in \cite{smith1990a} and the Lie algebra based methods presented in this paper (and $\Sigma_{2nd}$ in \cite[(55)]{barfoot2014associating}) are first order approximation methods (in terms of covariance), characterizing rotation uncertainty in the Lie algebra space results in a more accurate characterization of pose uncertainty. 
This is because Euler angle parameterization is a chart and does not cover the entire manifold whereas the Lie algebra inherently captures the group structure and can model any arbitrary group element. In addition, since the Lie algebra is a vector space~\cite{tapp2016matrix}, modeling the uncertainty as a Gaussian distribution is convenient and well-defined.

To demonstrate this, we performed an experiment similar to the one proposed in \cite{barfoot2014associating}. 
We generated a sequence of $N$ noisy pose transformations of the form $\mathbf{T}_{ab} = \operatorname{exp}(\boldsymbol{\xi}_{ab}^{\wedge}) \bar{\mathbf{T}}_{ab}$, with 
\begin{equation}
    \boldsymbol{\xi}_{ab} \sim \mathcal{N}(\mathbf{0}, \mathbf{\Sigma}_{ab}),
\end{equation}
\begin{equation}
    \mathbf{\Sigma}_{ab} = \operatorname{diag}([ 0.001\sigma_t, 1e^{-5}\sigma_t, 1e^{-5}, 1e^{-5}, 1e^{-5}, 0.003\sigma_r]),
\end{equation}
and
\begin{equation}
    \bar{\mathbf{T}}_{ab} = \left[ \begin{array}{cccc}
        1 & 0 & 0 & 1 \\
        0 & 1 & 0 & 0 \\
        0 & 0 & 1 & 0 \\
        0 & 0 & 0 & 1
    \end{array} \right],
\end{equation}
where $\sigma_t$ and $\sigma_r$ are scaling parameters for the translation and rotation noise respectively. 
In addition, each consecutive perturbation variable $\boldsymbol{\xi}_{ab}$ was drawn such that it was correlated with the 
perturbation variable before it with correlation coefficient $\rho$.

We then composed these transformations end-to-end and repeated the process to generate 10000 sample trajectories. 
These Monte-Carlo simulation results were then used to evaluate the uncertainty predicted by the SSC head-to-tail operation 
as well as the Lie-algebra based pose composition method derived in this section, both with and without correlation being taken 
into account. A top down view of one such experiment with $N=10$, $\sigma_t=5$, $\sigma_r=5$, and $\rho=0.4$ is shown in \figref{fig:compose_example}.

As expected, the Lie algebra based method significantly outperforms SSC head-to-tail because it takes into account the structure of the $\mathrm{SE}(3)$ group. In addition, it is easily seen that dropping the cross covariance terms 
results in an under-approximation of the true covariance, if positive correlation is  indeed present.
A more thorough investigation of this experiment is described in \secref{sec:eval:composition}.

In the next two sections, we provide a derivation of the inverse and relative pose operations on the Lie algebra, which as far as we know have not been previously published.

\section{The Pose Inverse Operation}
\label{sec:inverse}

The pose inverse operation corresponds to a change in reference frame. 
Given an uncertain pose distribution $\mathbf{T}_{ij}$ that represents the pose of coordinate frame $j$ with respect to frame $i$, we want to find the inverse $\mathbf{T}_{ji}$ that represents the pose of coordinate frame $i$ with respect to frame $j$. 
The distribution of an inverse pose can be derived as follows \cite{eade2017lie}:
\begin{align}
    \mathbf{T}_{ji} &= \mathbf{T}_{ij}^{-1} \nonumber\\
    &= \bar{\mathbf{T}}_{ij}^{-1} \operatorname{exp}(- \boldsymbol{\xi}_{ij}^{\wedge}) \label{eq:pose_inverse}\\
    &= \operatorname{exp}((-\mathrm{Ad}_{\bar{\mathbf{T}}_{ij}^{-1}}\boldsymbol{\xi}_{ij})^\wedge) \bar{\mathbf{T}}_{ij}^{-1} \nonumber\\
    &= \operatorname{exp}(\boldsymbol{\xi}'_{ij}) \bar{\mathbf{T}}_{ij}^{-1} \nonumber
\end{align}
where $\boldsymbol{\xi}'_{ij}$ is the original perturbation $\boldsymbol{\xi}_{ij}$ transformed by the negative adjoint of $\bar{\mathbf{T}}_{ij}^{-1}$. 
Due to the linearity of the adjoint operation, we can represent the inverse distribution $\mathbf{T}_{ji}$ with mean and covariance 
\begin{align}
    \bar{\mathbf{T}}_{ji} = \bar{\mathbf{T}}_{ij}^{-1}
\end{align}
and
\begin{align}
    \mathbf{\Sigma}_{ji} = \mathrm{Ad}_{\bar{\mathbf{T}}_{ij}^{-1}} \mathbf{\Sigma}_{ij} \mathrm{Ad}_{\bar{\mathbf{T}}_{ij}^{-1}}^\top.
\end{align}

\section{Finding the Relative Transformation Between Jointly Distributed Poses}
\label{sec:relative_pose}

This section combines the derivations from the previous two sections to formulate a first order method for characterizing the uncertainty of the relative pose operation or the coordinate-free equivalent to the SSC tail-to-tail operation described in Section \ref{sec:SSC:tail-to-tail}. As far as we can tell, this is the first time this operation has been published while using the Lie algebra to characterize uncertainty.

\subsection{Relative Pose Operation Derivation}

\begin{figure}[t]
    \centering
    \includegraphics[width=\columnwidth]{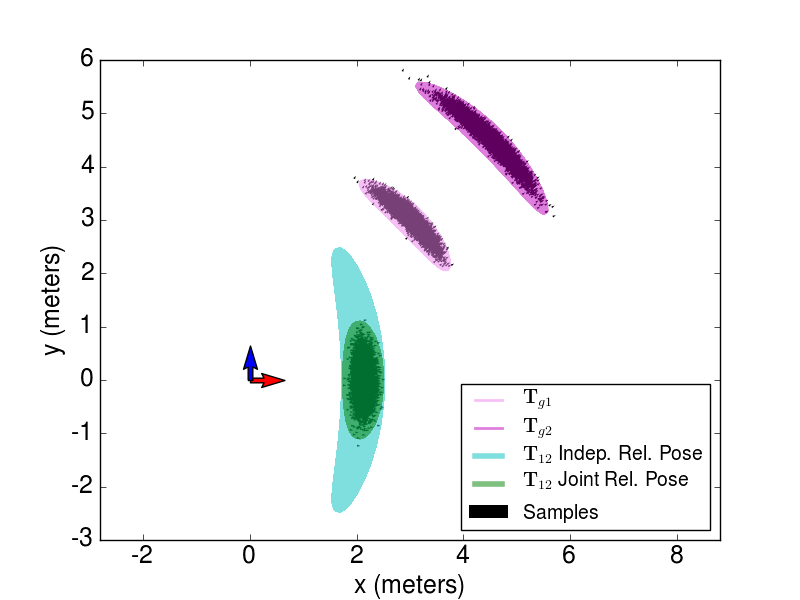}
    \caption{Dropping the correlation terms in \eqref{eq:pose_compound_cov} and \eqref{eq:relative_pose_cov} can lead to under/over-estimation of uncertainty. In this example, we used our proposed method to estimate the relative pose $\mathbf{T}_{12}$ between two correlated poses $\mathbf{T}_{g1}$ and $\mathbf{T}_{g2}$. A flattened view of the predicted 95\% likely uncertainty ellipsoids that result when correlation is ignored (\eqref{eq:relative_pose_cov} without the cross terms) and taken into account (\eqref{eq:relative_pose_cov} with all terms) are shown in cyan and green respectively. This plot was with $\alpha=1$.}
    \label{fig:between}
\end{figure}

\begin{figure}[t]
    \centering
    \includegraphics[width=\columnwidth]{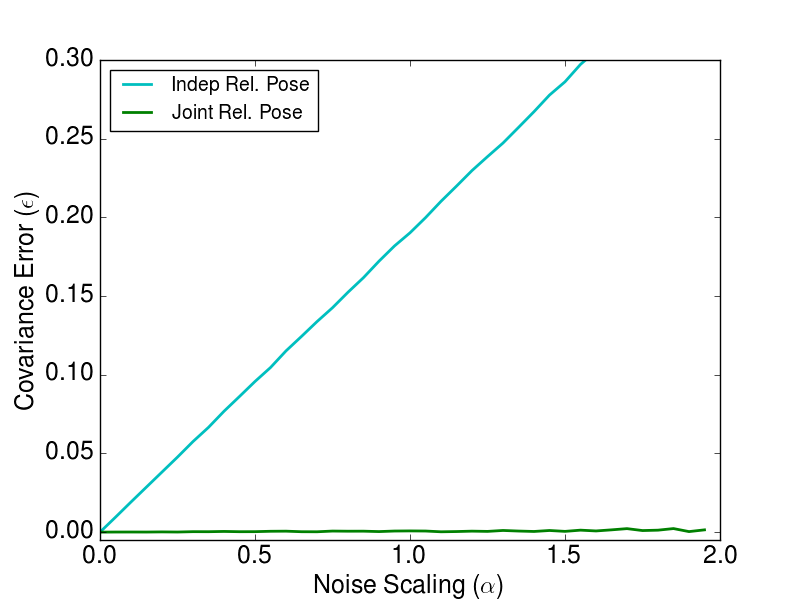}
    \caption{Covariance error comparison showing the importance of not ignoring correlation when propagating uncertainty.}
    \label{fig:cov_error}
\end{figure}

Given uncertain estimates of two coordinate frames with respect to a common base frame, our aim is estimate the relative transformation between the two poses. For example, given possibly 
correlated uncertain transformations $\mathbf{T}_{ij}$ and $\mathbf{T}_{ik}$ representing the poses of coordinate frames $j$ and $k$ with respect to frame $i$, we want to find the mean $\bar{\mathbf{T}}_{jk}$ and covariance $\mathbf{\Sigma}_{jk}$ that 
parameterize the pose uncertainty of frame $k$ with respect to frame $j$.

We start by assuming that the $\bar{\mathbf{T}}_{ij}$ and $\bar{\mathbf{T}}_{ik}$ are known and that the associated perturbations 
$\boldsymbol{\xi}_{ij}$ and $\boldsymbol{\xi}_{ik}$
are jointly correlated in the Lie algebra with known covariance
\begin{align}
    \mathbf{\Sigma} 
    &= 
    \left[ \begin{array}{cc}
        \mathbf{\Sigma}_{ij} & ~~~\mathbf{\Sigma}_{ij,ik}  \\
        ~~~\mathbf{\Sigma}_{ij,ik}^\top & \mathbf{\Sigma}_{ik}
    \end{array} \right]. \nonumber
\end{align} 
Under the standard group inverse and multiplication
actions, the following must hold:
\begin{equation}
    \mathbf{T}_{jk} = \mathbf{T}_{ij}^{-1} \mathbf{T}_{ik}. \label{eq:relative_pose}
\end{equation}
Expanding \eqref{eq:relative_pose} using the random variable definition in \eqref{eq:T_i} and the derivation in \eqref{eq:pose_inverse} results in
\begin{align}
    \operatorname{exp}(\boldsymbol{\xi}_{jk}^\wedge) \bar{\mathbf{T}}_{jk} 
    &= 
    \bar{\mathbf{T}}_{ij}^{-1}
    \operatorname{exp}(-\boldsymbol{\xi}_{ij}^\wedge) 
    \operatorname{exp}(\boldsymbol{\xi}_{ik}^\wedge) \bar{\mathbf{T}}_{ik} \\
    &= 
    \operatorname{exp}((-\mathrm{Ad}_{\bar{\mathbf{T}}_{ij}^{-1}}\boldsymbol{\xi}_{ij})^\wedge) \bar{\mathbf{T}}_{ij}^{-1}
    \operatorname{exp}(\boldsymbol{\xi}_{ik}^\wedge) \bar{\mathbf{T}}_{ik} \nonumber\\
    &= 
    \operatorname{exp}(\boldsymbol{\xi}_{ij}^{\prime\wedge})
    \operatorname{exp}(\boldsymbol{\xi}_{ik}^{\prime\wedge})
    \bar{\mathbf{T}}_{ij}^{-1}
     \bar{\mathbf{T}}_{ik}, \nonumber     
\end{align}
where $\boldsymbol{\xi}_{ij}^\prime = -\mathrm{Ad}_{\bar{\mathbf{T}}_{ij}^{-1}}\boldsymbol{\xi}_{ij}$ and $\boldsymbol{\xi}_{ik}^\prime = \mathrm{Ad}_{\bar{\mathbf{T}}_{ij}^{-1}}\boldsymbol{\xi}_{ik}$. 
Letting
\begin{equation}
    \bar{\mathbf{T}}_{jk} \triangleq \bar{\mathbf{T}}_{ij}^{-1} \bar{\mathbf{T}}_{ik}, \label{eq:relative_pose_mu}
\end{equation}
gives us
\begin{equation}
     \operatorname{exp}(\boldsymbol{\xi}_{jk}^\wedge) = \operatorname{exp}(\boldsymbol{\xi}_{ij}^{\prime\wedge})
    \operatorname{exp}(\boldsymbol{\xi}_{ik}^{\prime\wedge}).
\end{equation}
Expanding the BCH formula in a similar manner to
\eqref{eq:BCH_compound_pose} and taking the 
expectation as in \eqref{eq:2nd_order} results in the following up to first order:
\begin{align}
    E[\xib{jk}\xibt{jk}] \approx &E [ \xip{ij}\xipt{ij} ] + E[\xip{ik}\xipt{ik}] ~+  \label{eq:relative_pose_cov_expectation}\\       
    &+E[\xip{ij}\xipt{ik}] + E[\xip{ik}\xipt{ij}]. \nonumber
\end{align} 
Evaluating the expectations in \eqref{eq:relative_pose_cov_expectation} results in 
\begin{align}
    \mathbf{\Sigma}_{jk} \approx
    &\mathrm{Ad}_{\bar{\mathbf{T}}_{ij}^{-1}} \mathbf{\Sigma}_{ij} \mathrm{Ad}_{\bar{\mathbf{T}}_{ij}^{-1}}^\top +
    \mathrm{Ad}_{\bar{\mathbf{T}}_{ij}^{-1}} \mathbf{\Sigma}_{ik} \mathrm{Ad}_{\bar{\mathbf{T}}_{ij}^{-1}}^\top - \label{eq:relative_pose_cov}\\
    &+\mathrm{Ad}_{\bar{\mathbf{T}}_{ij}^{-1}} \mathbf{\Sigma}_{ij,ik} \mathrm{Ad}_{\bar{\mathbf{T}}_{ij}^{-1}}^\top -
    \mathrm{Ad}_{\bar{\mathbf{T}}_{ij}^{-1}} \mathbf{\Sigma}_{ij,ik}^\top \mathrm{Ad}_{\bar{\mathbf{T}}_{ij}^{-1}}^\top. \nonumber
\end{align}

Thus, uncertainty can be propagated through the relative pose function via
\eqref{eq:relative_pose_mu} and \eqref{eq:relative_pose_cov}. A summary of the uncertainty propagation methods derived in the last three sections is provided in \figref{fig:summary}.

\subsection{Ignoring Correlation Leads to Inconsistency}

 Ignoring correlation of the associated perturbation variables leads to under/over-estimation of uncertainty, depending on if the correlation is 
 positive or negative and if the operation being performed is pose composition or
 relative pose estimation.
\figref{fig:between} and \figref{fig:cov_error} show an example of this for the case of relative pose with positive correlation.

To create these plots, we generated $M=10000$ sets of two uncertain poses $\mathbf{T}^m_{g1}=\operatorname{exp}(\boldsymbol{\xi}_{g1}^{m\wedge}) \bar{\mathbf{T}}_{g1}$ and $\mathbf{T}_{g2}=\operatorname{exp}(\boldsymbol{\xi}_{g2}^{m\wedge}) \bar{\mathbf{T}}_{g2}$, with mean values 
\begin{equation}
    \bar{\mathbf{T}}_{g1} = \left[ \begin{array}{cccc}
        0.707107 & -0.707107  & 0      & 3 \\
        0.707107 &  0.707107  & 0      & 3 \\
       -0        &  0         & 1      & 0 \\
        0        &  0         & 0      & 1
    \end{array} \right]
\end{equation}  
and
\begin{equation}
    \bar{\mathbf{T}}_{g2} = \left[ \begin{array}{cccc}
        0.707107 & -0.707107  & 0      & 4.5 \\
        0.707107 &  0.707107  & 0      & 4.5 \\
       -0        &  0         & 1      & 0 \\
        0        &  0         & 0      & 1
    \end{array} \right],
\end{equation}
under the assumption that the perturbation variables ($\boldsymbol{\xi}^m_{g1}$) and ($\boldsymbol{\xi}^m_{g2}$) are jointly correlated in the Lie algebra with marginal covariance matrices
\begin{align}
    \mathbf{\Sigma}_{g1} &= \mathbf{\Sigma}_{g2}  \\
    &= \alpha \cdot \operatorname{diag}([0.005, 0.005, 1e-5, 1e-5, 1e-5, 0.006]),  \nonumber
\end{align}
and cross covariance
\begin{equation}
    \mathbf{\Sigma}_{g1,g2} = \alpha \cdot \operatorname{diag}([0.0005, 0.0005, 0, 0, 0, 0.005]),
\end{equation}
where $\alpha$ is a scaling parameter.
We then used the relative pose operation presented in this section to estimate $\mathbf{T}_{12}$ both with and without taking uncertainty into account.

In \figref{fig:cov_error}, we evaluate the estimated covariance
error for increasing values of $\alpha$ with respect to the Monte-Carlo simulation under the following covaraiance error metric 
\begin{equation}
    \epsilon \triangleq \sqrt{\operatorname{tr}\left( (\mathbf{\Sigma} - \mathbf{\Sigma}_{mc})^\top (\mathbf{\Sigma} - \mathbf{\Sigma}_{mc}) \right)}, \label{eq:cov_error}
\end{equation}
where 
\begin{equation}
\mathbf{\Sigma}_{mc} = \frac{1}{M}\sum_{m=1}^M \boldsymbol{\xi}_m \boldsymbol{\xi}_m^{\top},\label{eq:sample_mc_cov}
\end{equation}
with 
\begin{equation}
\mathbf{T}_m = (\operatorname{exp}(\boldsymbol{\xi}_{g1}^{m\wedge})\bar{\mathbf{T}}_{g1})^{-1}\operatorname{exp}(\boldsymbol{\xi}_{g2}^{m\wedge})\bar{\mathbf{T}}_{g2} \label{eq:sample_mc_mean}
\end{equation}
and 
\begin{equation}
\boldsymbol{\xi}_m = \operatorname{log}(\mathbf{T}_m \bar{\mathbf{T}}_{12}^{-1})^\vee.   
\label{eq:sample_mc_xi}
\end{equation}
\figref{fig:compose_example}, \figref{fig:between}, and \figref{fig:cov_error} show that ignoring correlation leads to inconsistent estimates of uncertainty.

\section{Converting to a Lie algebra Based Representation}
\label{sec:conversion}

Although uncertainty characterization in the Lie algebra is more accurate than coordinate based methods, in some cases an existing estimation algorithm may output pose estimates in an alternative parameterization. 
This section describes how to convert from an existing parameterization such as the SSC multi-variate Gaussian representation (defined in \eqref{eq:SSC_rep} and \eqref{eq:SSC_joint_rep}) to the proposed representation, as well as how to extract the mean and covariance of a jointly correlated set of poses from a MLE solution such as iSAM \cite{kaess2008isam} or g2o \cite{kummerle2011g2o} directly.  

\subsection{Converting from a Coordinate Based Representation}

\begin{figure}[t]
    \centering
    \includegraphics[width=\columnwidth]{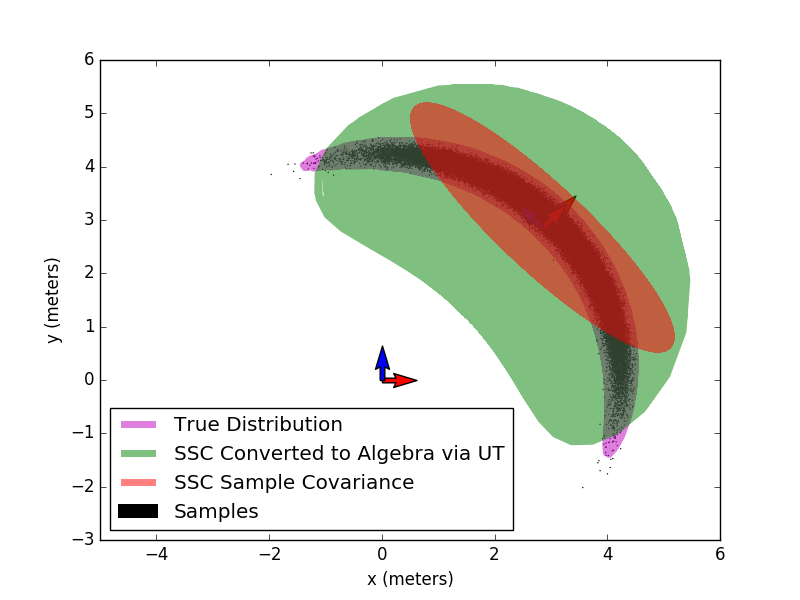}
    \caption{The Unscented Transform (UT) \cite{julier2002scaled} can be used to transform from an existing, coordinates based representation such as the one used in SSC \eqref{eq:SSC_rep} (shown in red) to the Lie algebra based representation (shown in green). However, the use of the SSC representation in the first 
    place results in a loss of information and the resulting representation (shown in green) over approximates the true underlying distribution (shown in pink). All uncertainty bounds
    shown are 95\% likely ellipsoids.}
    \label{fig:ut_conversion}
\end{figure}

Assuming we are given a mean parameter vector $\hat{\mathbf{x}}$ and associated covariance $\hat{\boldsymbol{\Sigma}}$ as defined in \eqref{eq:SSC_joint_rep}, to apply our proposed propagation methods, we must first transform $\hat{\mathbf{x}}$ and $\hat{\boldsymbol{\Sigma}}$ into into our proposed 
representation such that the means of the associated transformations are represented in the group space via $\bar{\mathbf{T}}_{1:n}$ and the perturbations are represented as multivariate Guassian random variables, $\boldsymbol{\xi}_{1:n}$,  with covariance $\boldsymbol{\Sigma}_{1:n}$ in the associated Lie algebra as defined in \eqref{eq:lie_joint_rep_cov} and \eqref{eq:lie_joint_rep_means}.
Converting from a coordinates based method for uncertainty characterization to the Lie algebra representation is a non-linear operation. 
As such, the unscented transform \cite{julier2002scaled} is a better choice for converting between representations than a first order linearization method. 

The unscented transform, first proposed by \citet{julier1997new} in 1997, takes an
input Gaussian distribution defined in one space and a non-linear function that maps the
input space to an output space and finds a Gaussian approximation for the transformed distribution. This is done by using the input mean and covariance to create a deterministic set $\mathcal{X}_{1:K}$ of $K = 2m + 1$ sigma points, where $m$ is the dimension of the input space. These sigma points are then passed through the non-linear
function and a weighted mean and sample covariance is used to find a Gaussian approximation to the output distribution. 

 Assuming we have a function $f : \mathbb{R}^d \mapsto \mathcal{G}$ that maps from a parameter vector, $\mathbf{x}_i$, to an element of the Lie group, $\mathcal{G}$, 
we can define a function $\ell_i$ 
that takes a perturbed parameter vector $\tilde{\mathbf{x}}_i$, centers it on the identity, and transforms it to the Lie algebra space
as follows:
\begin{equation}
    \ell_i(\tilde{\mathbf{x}}_i) = \operatorname{log}( f(\tilde{\mathbf{x}}_i) \cdot f(\hat{\mathbf{x}}_i)^{-1} ),
\end{equation}
where $\hat{\mathbf{x}}_i$ is the mean parameter vector associated with $\tilde{\mathbf{x}}_i$.

We can then use the standard unscented transform \cite[(12)]{julier1997new} to generate a set $\mathcal{X}_{1:K}$ of $2nd + 1$ sigma points from $\hat{\mathbf{x}}$ and $\hat{\boldsymbol{\Sigma}}$, where $n$ is the number of modeled Lie group elements. 
After which, the predicted Lie group representation for the jointly correlated poses can be obtained as follows:
\begin{equation}
    \bar{\mathbf{T}}_i = f(\hat{\mathbf{x}}_i)
\end{equation}
\begin{equation}
    \boldsymbol{\Sigma}_{1:n} = \sum_{k=1}^K \mathcal{W}_k \ell(\mathcal{X}_k) \ell(\mathcal{X}_k)^\top,
\end{equation}
where $\ell$ is a vectorized version of $\ell_i$ and $\mathcal{W}_k$ is the standard weight from \cite[(12)]{julier1997new}.
\figref{fig:ut_conversion} shows a result of this process.

 While, in many cases the use of the Unscented Transform may be the best that can be done, the use of the coordinates based representation (even as an intermediate representation) does lead to some loss of information. 
 This loss of information  can be avoided if we can directly represent the uncertainty in the proposed framework when extracting it from the prior estimation solution.

\subsection{Extracting Pose Uncertainty from a MLE Solution}
\label{sec:convert:cov_extraction}

State-of-the-Art Pose Graph SLAM solvers find a solution by estimating the set of robot poses that maximizes the likelihood of the observed measurements \cite{dellaert2006a, kaess2008isam, kaess2011isam2, rosen2016sesync, jmangelson-2019a}. 
Traditional, iterative non-linear solvers \cite{dellaert2006a, kaess2008isam, kaess2011isam2, kummerle2011g2o} do this by building up a measurement Jacobian, $A$, with columns that correspond to elements of the parameter vector $\mathbf{x}$ and with block rows that correspond to weighted, measurement residual error functions that minimize the error between predicted measurements and what was actually measured. 
At each iteration, this measurement Jacobian is used to form a linear least squares problem of the following form: 
\begin{equation}
    \hat{\mathbf{x}} = \underset{\mathbf{x}}{\operatorname{argmin}} \lVert A \mathbf{x} - \mathbf{b} \rVert^2,
\end{equation}
where $\mathbf{b}$ is the measurement vector not needed for the following derivation.
The algorithm alternates between solving this linear least squares optimization problem and relinearizing $A$ around $\hat{\mathbf{x}}$ until convergence \cite{dellaert2006a, kaess2008isam}. 
Algorithms such as those described in \cite{rosen2016sesync} and \cite{jmangelson-2019a} that provide a guarantee of global optimality formulate the problem slightly differently, but once a solution has been obtained, a matrix $A$ can still be formed by linearizing the cost around the current solution. 

After a solution has been reached, an estimate of the uncertainty of that solution can be found by using $A$ to form the information matrix $\mathcal{I}= A^\top A$ and using the non-zero elements of its Cholesky factorization $\mathcal{I}= R^\top R$ to calculate the necessary elements of the marginal covariance $\hat{\boldsymbol{\Sigma}}$ as detailed in \cite{kaess2009covariance}.
The trick to extracting this covariance with respect to $\boldsymbol{\xi}_{1:n}$ as opposed to $\hat{\mathbf{x}}$ is to make sure that the Jacobian $A$ used to form $\mathcal{I}$ is evaluated with respect to $\boldsymbol{\xi}_{1:n}$ as opposed to $\hat{\mathbf{x}}$.
This can be done by numerically evaluating the Jacobian and by perturbing $\boldsymbol{\xi}_{1:n}$ around $\mathbf{0}$ and propagating that perturbation to the linearization point by means of the exponential function as opposed to perturbing the parameters of  $\hat{\mathbf{x}}$ directly. 
Doing this enables the direct extraction of $\boldsymbol{\Sigma}_{1:n}$ and results in increased accuracy because $\boldsymbol{\xi}_{1:n}$ lies in a vector space, while $\hat{\mathbf{x}}$ does not.

\section{Evaluation}
\label{sec:eval}

This section evaluates the proposed uncertainty characterization method by performing two experiments. 
The first, preforms a parameter sweep over the experiment introduced in \secref{sec:pose_composition:comparison}. 
The second, extracts covariance information from the result of a Pose Graph SLAM algorithm and compares the predicted relative pose covariance with the sample covariance obtained from Monte Carlo.

\subsection{Compounding Correlated Odometry}
\label{sec:eval:composition}

\begin{figure}[t]
    \centering
    \includegraphics[width=\columnwidth]{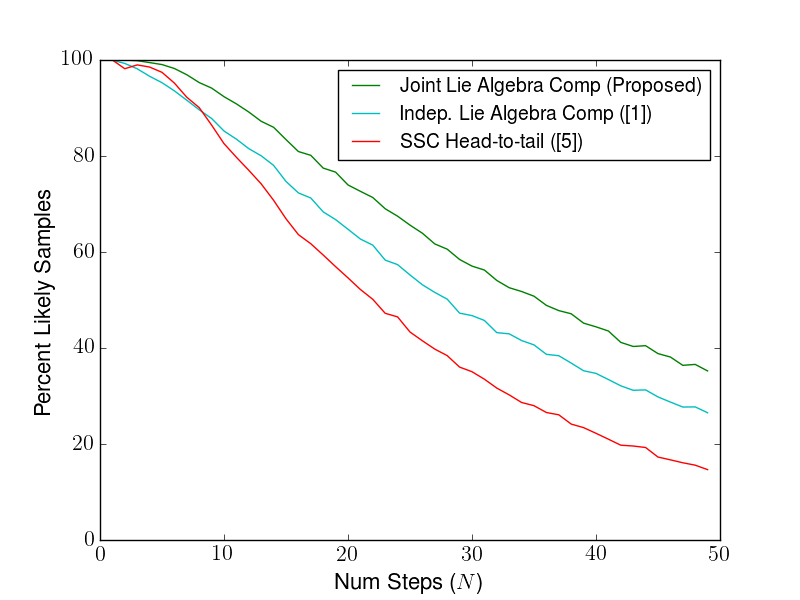}
    \caption{The percentage of final samples that fell within the 99.9\% likely covariance ellipsoid as a function of the number of 
    poses in the trajectory sequence. All methods drop off as the number of poses is increased, however our proposed method is consistently
    most accurate.}
    \label{fig:compose_num_steps}
\end{figure}
\begin{figure}[t]
    \centering
    \includegraphics[width=\columnwidth]{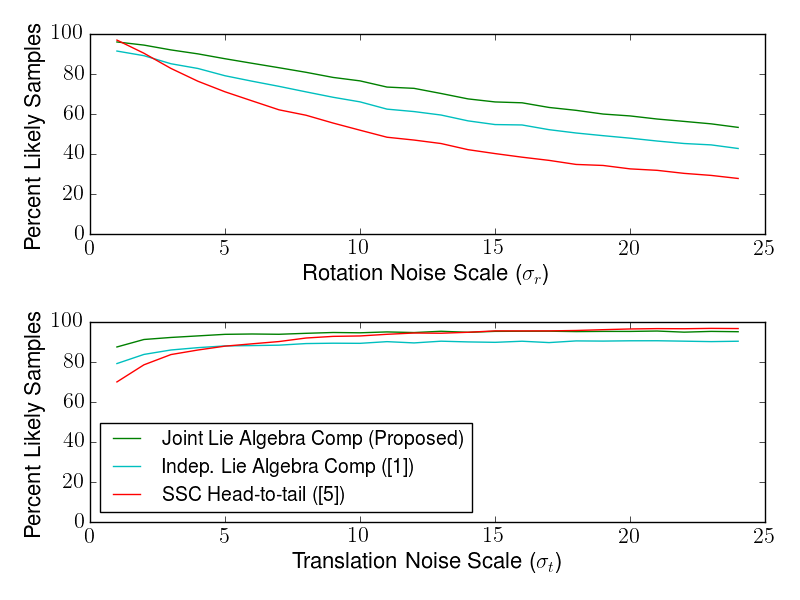}
    \caption{The percentage of final samples that fell within the 99.9\% likely covariance ellipsoid as a function of the rotation and translation noise. For the rotation noise sweep, translation noise was held constant at $\sigma_t=3$ and for the translation noise sweep, rotation noise was held constant at $\sigma_r=3$. Note that increases in rotation noise have the largest negative effect. }
    \label{fig:compose_noise}
\end{figure}

The exact accuracy of our proposed uncertainty characterization method is dependent on a variety of parameters including the number of poses compounded end-to-end and the rotation and translation noise. 
We explore this dependence by performing a parameter sweep across each of these parameters based on the experiment introduced in \secref{sec:pose_composition:comparison}.

\begin{figure*}[!t]%
    \centering%
    \subfloat[Pose Pair Correlation - X\label{fig:x_corr_50}]{%
       \includegraphics[width=0.33\textwidth,trim={1cm 0 1cm 0},clip]{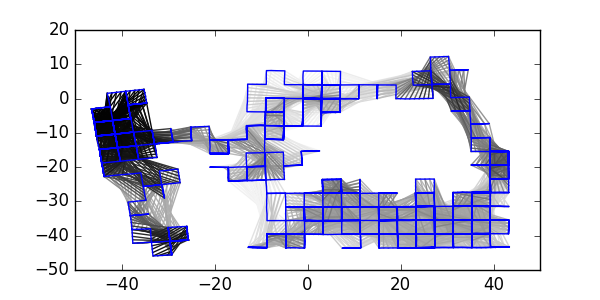}
    }
    \subfloat[Pose Pair Correlation - Y\label{fig:y_corr_50}]{%
       \includegraphics[width=0.33\textwidth,trim={1cm 0 1cm 0},clip]{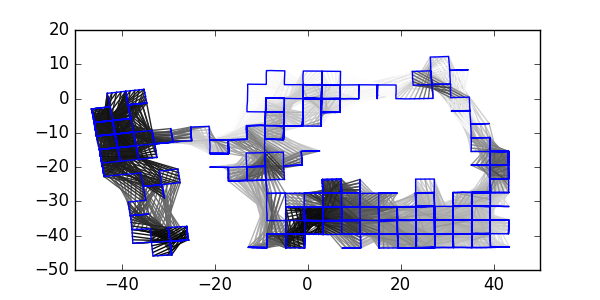}
    }
    \subfloat[Pose Pair Correlation - Heading\label{fig:t_corr_50}]{%
       \includegraphics[width=0.33\textwidth,trim={1cm 0 1cm 0},clip]{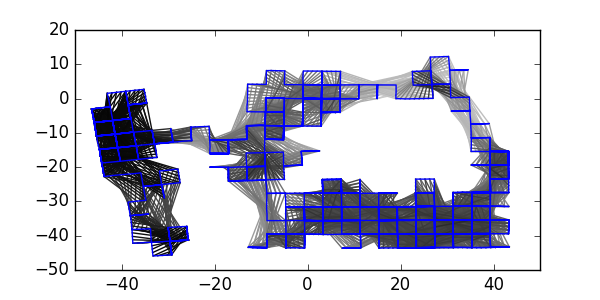}
    }
    \hfill
    \subfloat[Proposed Lie Algebra Relative Pose Cov. Error\label{fig:joint_error_50}]{%
       \includegraphics[width=0.4\textwidth,trim={1cm 0 1cm 0},clip]{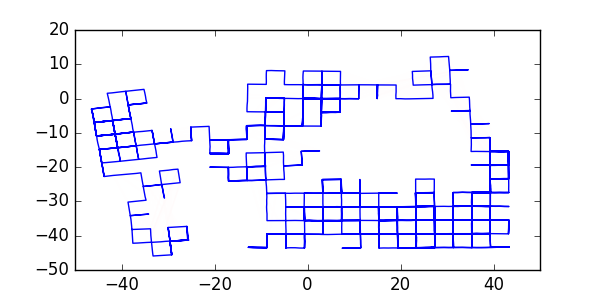}
    }
    \subfloat[Proposed Relative Pose Cov. Error (Ignoring Correlation)\label{fig:indep_error_50}]{%
       \includegraphics[width=0.4\textwidth,trim={1cm 0 1cm 0},clip]{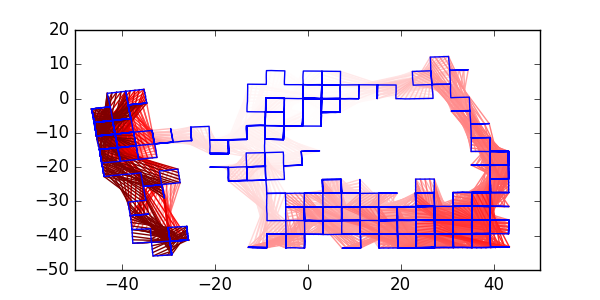}
    }\hfill
    \subfloat[Proposed Lie Algebra Relative Pose Normalized Cov. Error\label{fig:norm_joint_error_50}]{%
       \includegraphics[width=0.4\textwidth,trim={1cm 0 1cm 0},clip]{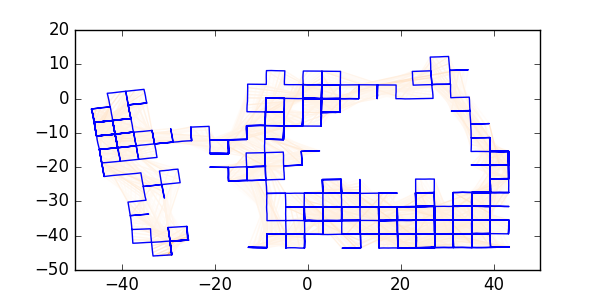}
    }
    \subfloat[SSC Relative Pose Normalized Cov. Error\label{fig:norm_ssc_error_50}]{%
       \includegraphics[width=0.4\textwidth,trim={1cm 0 1cm 0},clip]{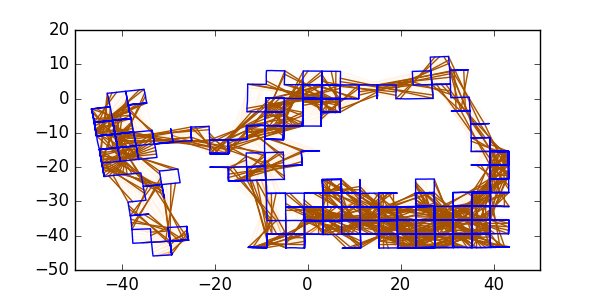}
    }\hfill
    \caption{A visualization of the correlation and covariance error for 3450 pose pairs with an offset of 50 nodes extracted from a solution of the Manhattan3500 dataset \cite{olson2006fast}. \protect\subref{fig:x_corr_50}, \protect\subref{fig:y_corr_50}, and \protect\subref{fig:t_corr_50} show the relative poses colored by the correlation coefficient of the $x$, $y$, and $\theta$ dimensions respectively. White corresponds to a correlation coefficient of 0 and black to a coefficient of 1. \protect\subref{fig:joint_error_50} and \protect\subref{fig:indep_error_50} show the covariance error with respect to Monte Carlo for relative pose estimation in the Lie algebra when correlation is and is not taken into account. Dark red corresponds to a covariance error of at least 2 standard deviations above the mean (with respect to when correlation is ignored), while white corresponds to a covariance error of 0. \protect\subref{fig:norm_joint_error_50} and \protect\subref{fig:norm_ssc_error_50} show the normalized covariance error with respect to Monte Carlo for the proposed method and SSC \cite{smith1990a}. In this case, dark orange corresponds to a covariance error of at least 2 standard deviations above the mean (with respect to SSC relative pose extraction), while white corresponds to a covariance error of 0.}
    \label{fig:slam_rp_error_50}
\end{figure*}

We began by varying the number of poses in the trajectory ($N$) and held both noise scale parameters fixed at  $\sigma_t = \sigma_r = 3$. 
The results are shown in \figref{fig:compose_num_steps}. 
As the number of steps increases, the accuracy of the final estimate drops off, this is because each of the methods are only characterizing uncertainty up-to first order and the lost higher order information builds up as the number of poses increases. 

\begin{table}[t!]
  \addtolength{\tabcolsep}{-5pt}
  \centering
  \caption{Summary of covariance error statistics for proposed Lie algebra relative pose estimation when correlation is taken into account and ignored for a total
  of 44425 pose pairs extracted from a solution to the Manhattan3500 dataset \cite{olson2006fast}, with pose offsets
  ranging from 5 to 500 nodes. }
  \scalebox{0.88}{  
  \begin{tabular}{lcc}
    \toprule
    Method & Covariance Error Mean & ~~Covariance Error Std. Dev. \\
    \midrule
    Proposed Lie algebra Rel. Pose &  0.00675104 $\pm~ 2.2e^{-4}$ &  0.0461455\\ 
    Proposed (Ignoring Correlation) & 2.05667 $\pm~ 1.0e^{-2}$ & 2.12371 \\ 
    \bottomrule
  \end{tabular}
}
  \label{table:cov_error}
\end{table}

For the noise parameter experiments, we varied the noise scale parameters $\sigma_r$ and $\sigma_t$ in turn, while keeping the number of poses fixed at $N=10$ and the non-varying noise parameter fixed at $3$. 
The results are shown in \figref{fig:compose_noise}. 
Rotation noise has a much more significant effect, again this is because of the lost higher-order information (either in terms of the higher order terms of the BCH formula or through linearization for SSC). 
Both Lie group methods consistently outperform SSC except when rotation noise  is very low and translation noise is very high. 
This is because when rotation error is very low, the non-linearity of the transformation becomes almost negligible and because the increased translation error causes the covariance ellipsoid to increase to the point where it includes the final robot position.
In addition the joint composition method in the Lie algebra is consistently more accurate than when correlation is ignored. 
For all three parameter sweeps, the induced correlation was held fixed at $\rho=0.4$.

\subsection{Extracting Relative Pose From a SLAM Solution}
\label{sec:eval:relative}

\begin{table}[t!]
  \addtolength{\tabcolsep}{-5pt}
  \centering
  \caption{Summary of normalized covariance error statistics for proposed and SSC \cite{smith1990a} relative pose estimation for a total of 44425 pose pairs extracted from a solution to the Manhattan3500 dataset \cite{olson2006fast}, with pose offsets
  ranging from 5 to 500 nodes. }
  \scalebox{0.88}{  
  \begin{tabular}{lcc}
    \toprule
    Method & Normalized Cov. Error Mean & ~~Normalized Cov. Error Std. Dev. \\
    \midrule
    Proposed Rel. Pose &  0.0493121 $\pm~ 1.7e^{-4}$ & 0.0353072\\ 
    SSC Tail-to-Tail & 0.28778 $\pm~ 2.0e^{-3}$ & 0.411625 \\ 
    \bottomrule
  \end{tabular}
}
  \label{table:norm_cov_error}
\end{table}

\begin{figure*}[!t]%
    \centering%
    \subfloat[Pose Pair Correlation - X\label{fig:x_corr_10}]{%
       \includegraphics[width=0.33\textwidth,trim={1cm 0 1cm 0},clip]{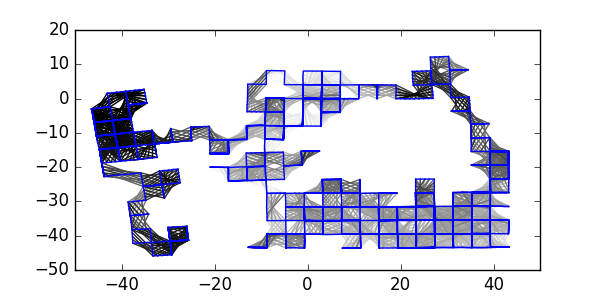}
    }
    \subfloat[Pose Pair Correlation - Y\label{fig:y_corr_10}]{%
       \includegraphics[width=0.33\textwidth,trim={1cm 0 1cm 0},clip]{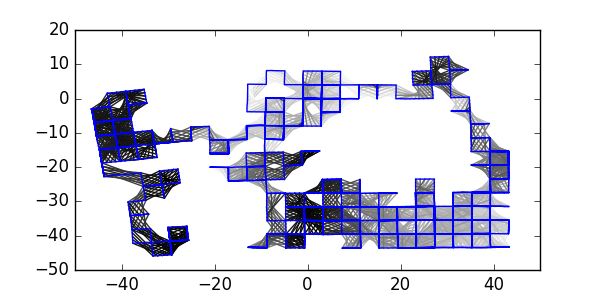}
    }
    \subfloat[Pose Pair Correlation - Heading\label{fig:t_corr_10}]{%
       \includegraphics[width=0.33\textwidth,trim={1cm 0 1cm 0},clip]{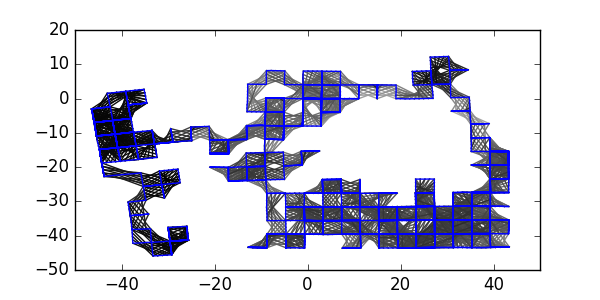}
    }
    \hfill
    \subfloat[Proposed Lie Algebra Relative Pose Cov. Error\label{fig:joint_error_10}]{%
       \includegraphics[width=0.39\textwidth,trim={1cm 0 1cm 0},clip]{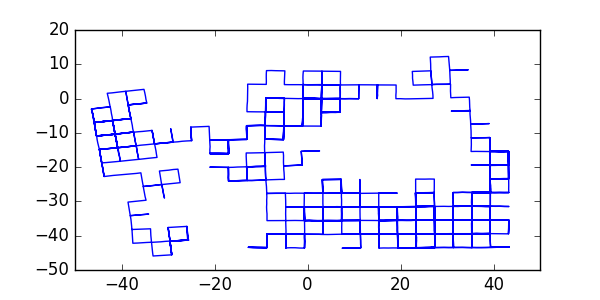}
    }
    \subfloat[Proposed Relative Pose Cov. Error (Ignoring Correlation)\label{fig:indep_error_10}]{%
       \includegraphics[width=0.39\textwidth,trim={1cm 0 1cm 0},clip]{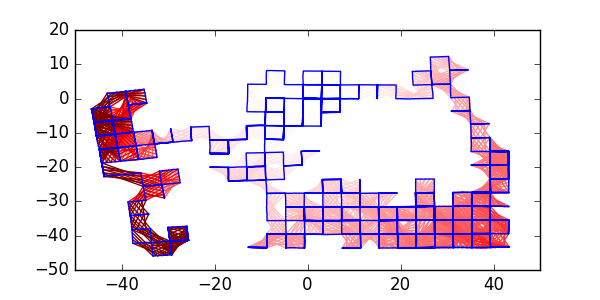}
    }\hfill
        \subfloat[Proposed Lie Algebra Relative Pose Normalized Cov. Error\label{fig:norm_joint_error_10}]{%
       \includegraphics[width=0.39\textwidth,trim={1cm 0 1cm 0},clip]{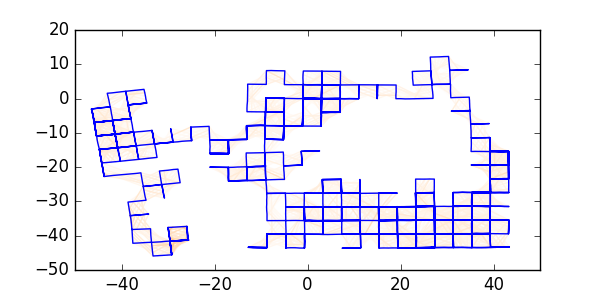}
    }
    \subfloat[SSC Relative Pose Normalized Cov. Error\label{fig:norm_ssc_error_10}]{%
       \includegraphics[width=0.39\textwidth,trim={1cm 0 1cm 0},clip]{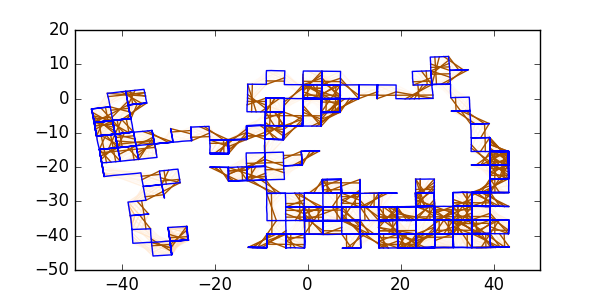}
    }\hfill
  \caption{A visualization of the correlation and covariance error for 3490 pose pairs with an offset of 10 nodes extracted from a solution of the Manhattan3500 dataset \cite{olson2006fast}. The color schemes match those of \figref{fig:slam_rp_error_50}.}
  \label{fig:slam_rp_error_10}
\end{figure*}

\begin{figure*}[!t]%
    \centering%
    \subfloat[Pose Pair Correlation - X\label{fig:x_corr_100}]{%
       \includegraphics[width=0.33\textwidth,trim={1cm 0 1cm 0},clip]{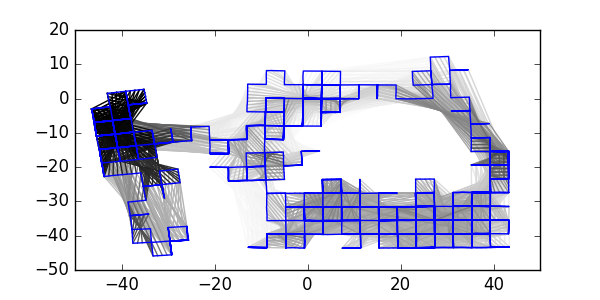}
    }
    \subfloat[Pose Pair Correlation - Y\label{fig:y_corr_100}]{%
       \includegraphics[width=0.33\textwidth,trim={1cm 0 1cm 0},clip]{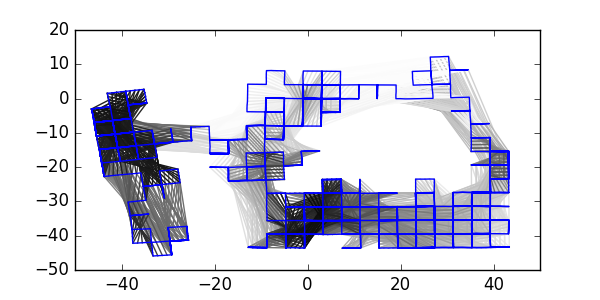}
    }
    \subfloat[Pose Pair Correlation - Heading\label{fig:t_corr_100}]{%
       \includegraphics[width=0.33\textwidth,trim={1cm 0 1cm 0},clip]{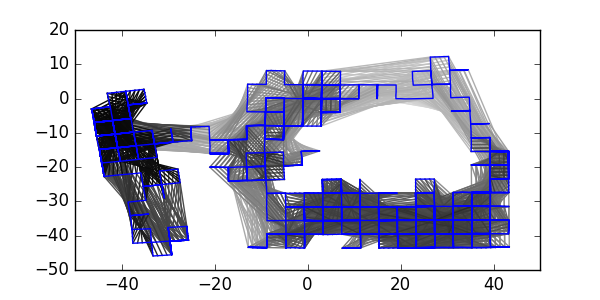}
    }
    \hfill
    \subfloat[Proposed Lie Algebra Relative Pose Cov. Error\label{fig:joint_error_100}]{%
       \includegraphics[width=0.39\textwidth,trim={1cm 0 1cm 0},clip]{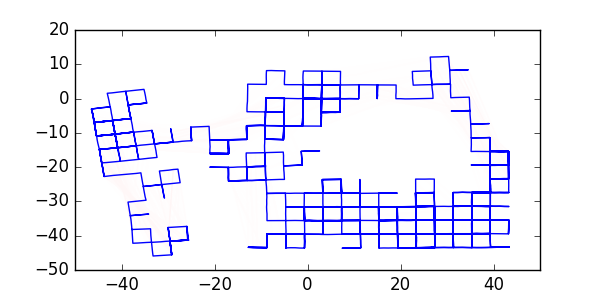}
    }
    \subfloat[Proposed Relative Pose Cov. Error (Ignoring Correlation)\label{fig:indep_error_100}]{%
       \includegraphics[width=0.39\textwidth,trim={1cm 0 1cm 0},clip]{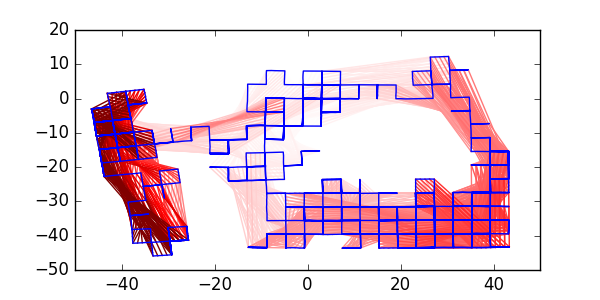}
    }\hfill
        \subfloat[Proposed Lie Algebra Relative Pose Normalized Cov. Error\label{fig:norm_joint_error_100}]{%
       \includegraphics[width=0.39\textwidth,trim={1cm 0 1cm 0},clip]{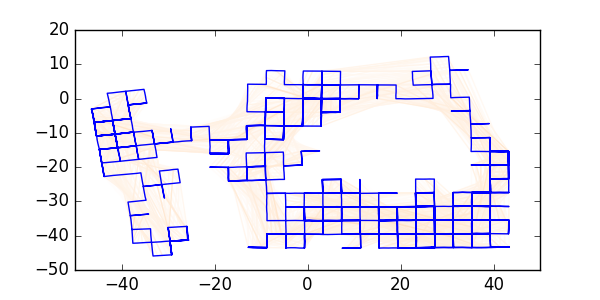}
    }
    \subfloat[SSC Relative Pose Normalized Cov. Error\label{fig:norm_ssc_error_100}]{%
       \includegraphics[width=0.39\textwidth,trim={1cm 0 1cm 0},clip]{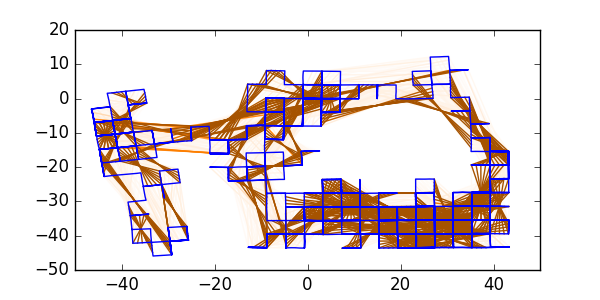}
    }\hfill
  \caption{A visualization of the correlation and covariance error for 3400 pose pairs with an offset of 100 nodes extracted from a solution of the Manhattan3500 dataset \cite{olson2006fast}. The color schemes match those of \figref{fig:slam_rp_error_50}.}
  \label{fig:slam_rp_error_100}
\end{figure*}

\begin{figure*}[!t]%
    \centering%
    \subfloat[Pose Pair Correlation - X\label{fig:x_corr_200}]{%
       \includegraphics[width=0.33\textwidth,trim={1cm 0 1cm 0},clip]{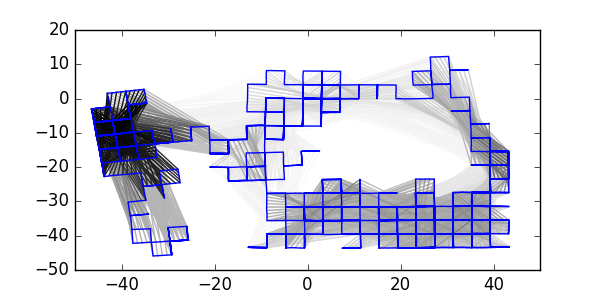}
    }
    \subfloat[Pose Pair Correlation - Y\label{fig:y_corr_200}]{%
       \includegraphics[width=0.33\textwidth,trim={1cm 0 1cm 0},clip]{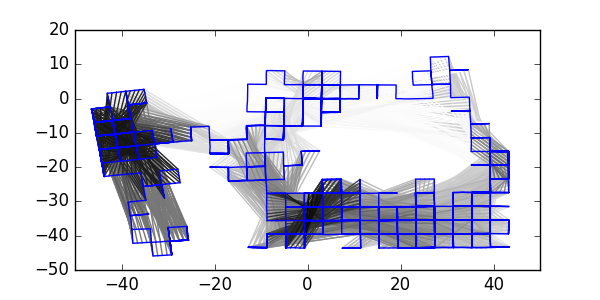}
    }
    \subfloat[Pose Pair Correlation - Heading\label{fig:t_corr_200}]{%
       \includegraphics[width=0.33\textwidth,trim={1cm 0 1cm 0},clip]{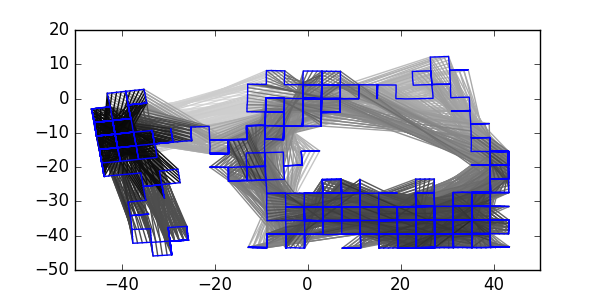}
    }
    \hfill
    \subfloat[Proposed Lie Algebra Relative Pose Cov. Error\label{fig:joint_error_200}]{%
       \includegraphics[width=0.39\textwidth,trim={1cm 0 1cm 0},clip]{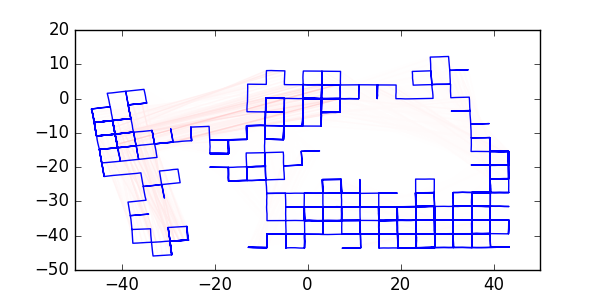}
    }
    \subfloat[Proposed Relative Pose Cov. Error (Ignoring Correlation)\label{fig:indep_error_200}]{%
       \includegraphics[width=0.39\textwidth,trim={1cm 0 1cm 0},clip]{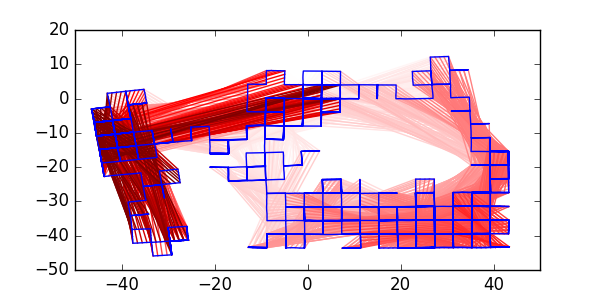}
    }\hfill
        \subfloat[Proposed Lie Algebra Relative Pose Normalized Cov. Error\label{fig:norm_joint_error_200}]{%
       \includegraphics[width=0.39\textwidth,trim={1cm 0 1cm 0},clip]{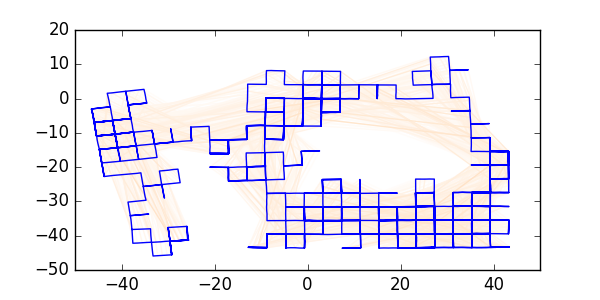}
    }
    \subfloat[SSC Relative Pose Normalized Cov. Error\label{fig:norm_ssc_error_200}]{%
       \includegraphics[width=0.39\textwidth,trim={1cm 0 1cm 0},clip]{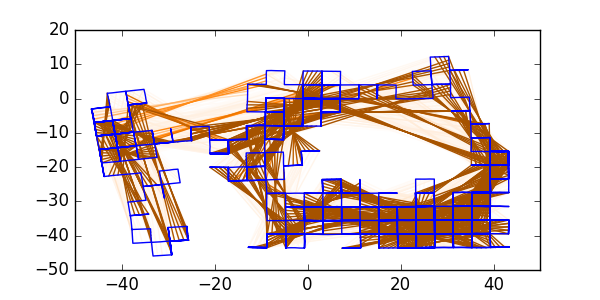}
    }\hfill
  \caption{A visualization of the correlation and covariance error for 3300 pose pairs with an offset of 200 nodes extracted from a solution of the Manhattan3500 dataset \cite{olson2006fast}. The color schemes match those of \figref{fig:slam_rp_error_50}.}
  \label{fig:slam_rp_error_200}
\end{figure*}

\begin{figure*}[!t]%
    \centering%
    \subfloat[Pose Pair Correlation - X\label{fig:x_corr_500}]{%
       \includegraphics[width=0.33\textwidth,trim={1cm 0 1cm 0},clip]{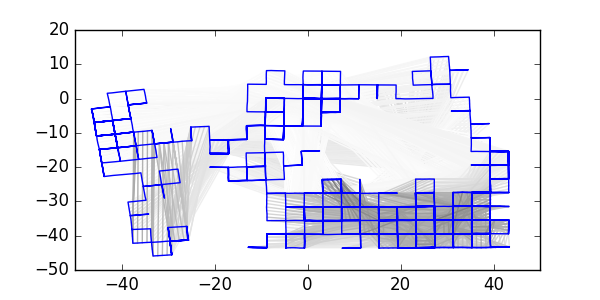}
    }
    \subfloat[Pose Pair Correlation - Y\label{fig:y_corr_500}]{%
       \includegraphics[width=0.33\textwidth,trim={1cm 0 1cm 0},clip]{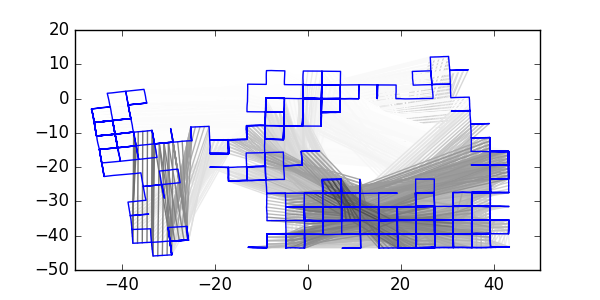}
    }
    \subfloat[Pose Pair Correlation - Heading\label{fig:t_corr_500}]{%
       \includegraphics[width=0.33\textwidth,trim={1cm 0 1cm 0},clip]{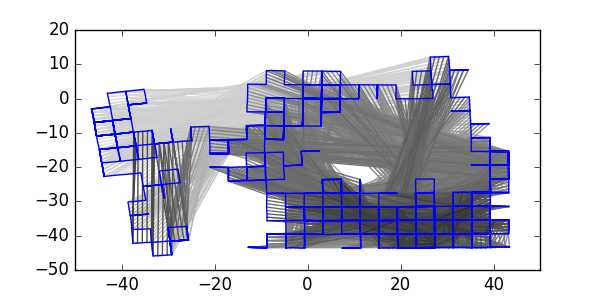}
    }
    \hfill
    \subfloat[Proposed Lie Algebra Relative Pose Cov. Error\label{fig:joint_error_500}]{%
       \includegraphics[width=0.39\textwidth,trim={1cm 0 1cm 0},clip]{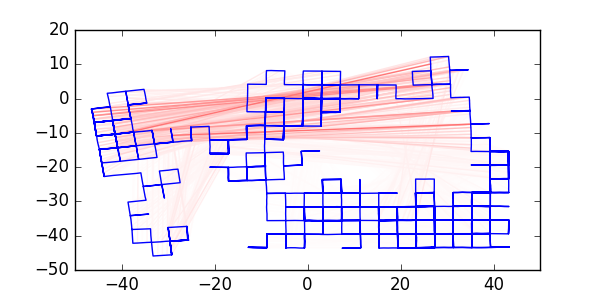}
    }
    \subfloat[Proposed Relative Pose Cov. Error (Ignoring Correlation)\label{fig:indep_error_500}]{%
       \includegraphics[width=0.39\textwidth,trim={1cm 0 1cm 0},clip]{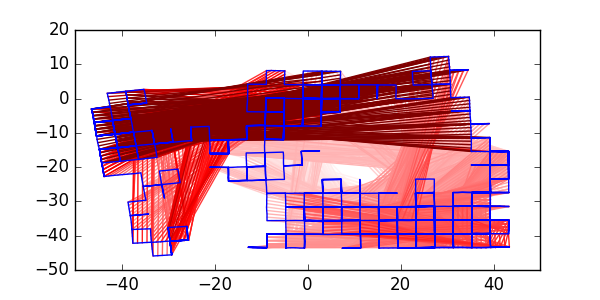}
    }\hfill
        \subfloat[Proposed Lie Algebra Relative Pose Normalized Cov. Error\label{fig:norm_joint_error_500}]{%
       \includegraphics[width=0.39\textwidth,trim={1cm 0 1cm 0},clip]{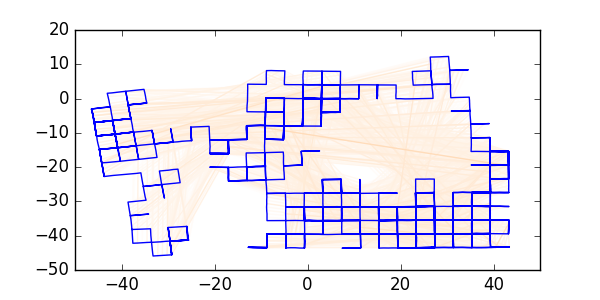}
    }
    \subfloat[SSC Relative Pose Normalized Cov. Error\label{fig:norm_ssc_error_500}]{%
       \includegraphics[width=0.39\textwidth,trim={1cm 0 1cm 0},clip]{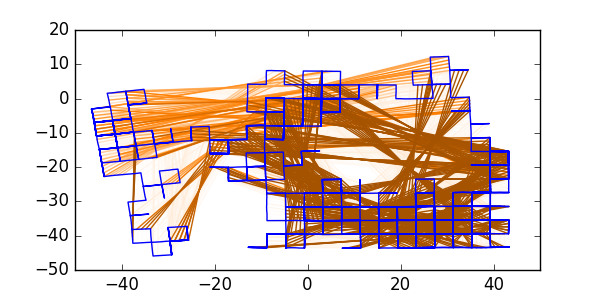}
    }\hfill
  \caption{A visualization of the correlation and covariance error for 3000 pose pairs with an offset of 500 nodes extracted from a solution of the Manhattan3500 dataset \cite{olson2006fast}. The color schemes match those of \figref{fig:slam_rp_error_50}.}
  \label{fig:slam_rp_error_500}
\end{figure*}


To evaluate relative pose extraction, we used iSAM \cite{kaess2008isam} to find a solution to the Manhattan 3500 dataset \cite{olson2006fast}. We then
extracted joint mean and covariance as described in \secref{sec:convert:cov_extraction} for pairs of poses at offsets varying from 5 to 50 in increments of 5 as well as for offsets of 100, 200, and 500 nodes. A visualization of the extracted pose pairs and the correlation between them for offsets of 50 are shown in \figref{fig:slam_rp_error_50} (a-c). Equivalent visualizations for offsets of 10, 100, 200, and 500 are shown in \figref{fig:slam_rp_error_10}, \figref{fig:slam_rp_error_100}, \figref{fig:slam_rp_error_200}, and \figref{fig:slam_rp_error_500}.

To investigate the importance of taking into account correlation when estimating relative pose, we performed a Monte Carlo simulation to estimate the true relative pose covariance as described in \eqref{eq:sample_mc_cov} - \eqref{eq:sample_mc_xi}, where $\boldsymbol{\xi}_{g1}^{m}$ and $\boldsymbol{\xi}_{g2}^{m}$ are the perturbation variables sampled from the extracted joint covariance and $\bar{\mathbf{T}}_{g1}$ and $\bar{\mathbf{T}}_{g2}$ are the mean values extracted from the iSAM solution. We then used the metric defined in \eqref{eq:cov_error} to evaluate the
covariance error of our method proposed in \eqref{eq:relative_pose_cov} when taking into account correlation (using all four terms in \eqref{eq:relative_pose_cov}) or ignoring correlation (using only the first two terms in \eqref{eq:relative_pose_cov}). Summary statistics are shown in Table \ref{table:cov_error} and visualizations of the error for offsets of 50, 10, 100, 200, and 500 are shown in (d-e) of \figref{fig:slam_rp_error_50}, \figref{fig:slam_rp_error_10}, \figref{fig:slam_rp_error_100}, \figref{fig:slam_rp_error_200}, and \figref{fig:slam_rp_error_500}.

The results in Table \ref{table:cov_error} show that ignoring correlation can lead to a covariance error more than 3 orders of magnitude higher than if correlation is taken into account. 

To compare our proposed method to SSC \cite{smith1990a}, we performed a similar experiment except that the Monte Carlo "groundtruth" covariance for SSC was derived by taking the sample relative poses from the previous experiment and converting them to the parameter vector format described in \eqref{eq:SSC_rep} and taking the sample covariance. To 
fairly compare our proposed method and SSC, the covariance matrices were normalized by the Frobenious norm of the Monte
Carlo covariance matrix before evaluating the covariance error as defined in \eqref{eq:cov_error}. Summary statistics
of the experiment are shown in Table \ref{table:norm_cov_error} and visualizations of the error for offsets of 50, 10, 100, 200, and 500 are shown in (f-g) of \figref{fig:slam_rp_error_50}, \figref{fig:slam_rp_error_10}, \figref{fig:slam_rp_error_100}, \figref{fig:slam_rp_error_200}, and \figref{fig:slam_rp_error_500}.

It should be noted that this is not a perfect comparison because the Monte Carlo covariance to which SSC is being compared is a multivariate Gaussian fit to a set of parameter vectors
that cannot be accurately modeled as a Gaussian. However, the results do show that even when modeling the true error in the format assumed by the SSC representation, our proposed Lie algebra
based method results in an order of magnitude lower covariance error than SSC (see Table \ref{table:norm_cov_error}).

\section{Library Implementation}
\label{sec:library}

We have released an open source C++ library implementation of our method. It can be downloaded at: https://bitbucket.org/jmangelson/lie. We have tried to 
design it to be simple, intuitive, and easily extendable. 

\subsection{Creating Known and Uncertain $\mathrm{SE}(3)$ Objects}

Creating both known and uncertain $\mathrm{SE}(3)$ objects is syntactically easy.
After importing the library a variety of constructors can be used to create known $\mathrm{SE}(3)$ transformations:
\begin{verbatim}
#include <lie/se3.hpp>

// Via rotation and translation 
Eigen::MatrixXd R(3,3); // 3x3 matrix 
Eigen::VectorXd t(3); // 3x1 vector
Lie::SE3 T_12(R, t);    
    
// Via translation and Euler angle params
Lie::SE3 T_23(x, y, z, theta, phi, psi);
\end{verbatim}

Independent and jointly distributed sets of unknown poses can also be created  by passing in  a mean value or a tuple of mean values and a covariance matrix.
\begin{verbatim}
Lie::SE3 T_ab_mu(R_ab, t_ab);  
Eigen::MatrixXd Sigma_ab(6,6);
auto T_ab_uncertain = 
  Lie::make_uncertain_state(
    T_ab_mu, Sigma_ab);
   
Lie::SE3 T_ac_mu(R_ac, t_ac);  
Eigen::MatrixXd Sigma_Joint(12,12);
auto refs = 
  Lie::make_uncertain_state(
    std::make_tuple(T_ab_mu, T_ac_mu), 
    Sigma_Joint);
auto T_ab_uncertain = std::get<0>(refs);
auto T_ac_uncertain = std::get<1>(refs);
\end{verbatim}

The \texttt{Lie::make\_uncertain\_state} function returns either a single $\mathrm{SE}(3)$ reference object or a tuple of such objects that can  then be used to perform operations.

\subsection{Performing Operations}

The pose composition, inverse, and relative pose operations  can be applied interchangeably regardless of whether or not  individual $\mathrm{SE}(3)$ objects are known or uncertain  via the \texttt{Lie::compose}, \texttt{inverse}, and \texttt{Lie::between} functions respectively. 
The compiler automatically determines whether or not the individual  objects are known/unknown or independent/jointly distributed  and apply the appropriate formulation. 

\begin{verbatim}
// Composing known poses
auto T_13 = Lie::compose(T_12, T_23);
  
// Invert both known and uncertain poses
auto T_21 = T_12.inverse();
auto T_ba_uncertain = 
  T_ab_uncertain.inverse();
  
// Calculating relative pose
auto T_bc_uncertain = Lie::between(
  T_ab_uncertain, T_ac_uncertain);
auto T_bc_uncertain2 = Lie::between(
  T_ab_known, T_ac_uncertain);
\end{verbatim}

Each function returns a new (from then on assumed independent) $\mathrm{SE}(3)$ reference object that can be used for additional operations as necessary.

\subsection{Additional Lie groups}

In addition, because the uncertainty propagation method we  present simply uses the properties of Lie groups, it can easily be applied to other Lie group types. 
We have also implemented the \texttt{Lie::SO3}, \texttt{Lie::SE2}, and \texttt{Lie::SO2} classes for the $\mathrm{SO}(3)$, $\mathrm{SE}(2)$, and $\mathrm{SO}(2)$ Lie groups and plan to extend it to other Lie group types as needed. 
\section{Conclusion}
\label{sec:conclusion}

Recent interest has shown that pose uncertainty  characterization through the use of the Lie algebra leads to more accurate uncertainty propagation. 
However, recent work assumes that individual poses are independent from one another and have primarily focused on pose composition, ignoring the equally important inverse and relative pose operations. 

This paper describes how to represent multiple jointly correlated poses while using the Lie algebra to characterize uncertainty. 
We also derive the equivalent of the Smith, Self, and Cheeseman \cite{smith1990a} pose composition, pose inverse, and relative pose operations when using the proposed framework. Finally, we have released an open source C++ library implementation of our method. 

The proposed methods can be used to increase the  accuracy of data association consistency checks when extracting pose uncertainty information from a pose graph \ac{SLAM} solution \cite{jmangelson-2018a}. It can also be used to compose odometry measurements that are potentially correlated with one another such as can be the case in the presence of wheel slip. By accurately 
modeling pose uncertainty, the proposed method increases the robustness and reliability of autonomous navigation.

\balance
\bibliographystyle{style/IEEEtranN}
{\footnotesize
\bibliography{references/IEEEabrv,references/strings-short,%
    references/library}}

\begin{IEEEbiography}[{\includegraphics[width=1in,height=1.25in,clip,keepaspectratio]{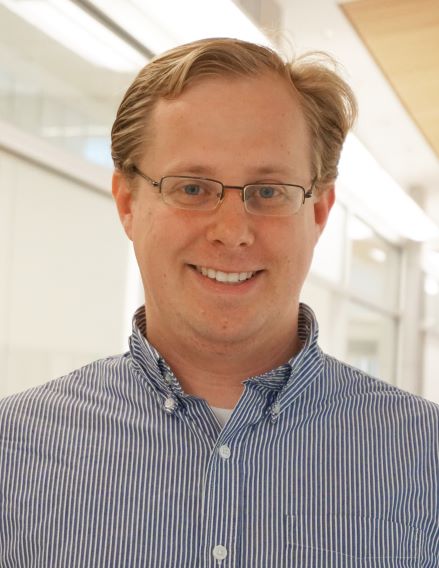}}]{Joshua Mangelson}

received the B.S. degree in electrical engineering from Brigham Young University, Provo, UT, USA in 2014 and the M.S. degree in robotics from the University of Michigan, Ann Arbor, MI, USA in 2016. He is currently a Ph.D. candidate in robotics at the University of Michigan with expected completion in April 2019. His research interests lie in the development of navigation, mapping, perception, and planning algorithms with mathematical and performance guarantees that enable the design of reliable field robotic systems for operation in unstructured environments. He is especially interested in the development of large-scale multi-agent teams for autonomous inspection of underwater structures. He is the recipient of the IEEE ICRA Best Multi-Robot Paper Award and the IEEE OCEANS Best Poster Award both in 2018. 
\end{IEEEbiography}

\vspace{-2 cm}

\begin{IEEEbiography}[{\includegraphics[width=1in,height=1.25in,clip,trim={4.7cm 0 5.7cm 0},keepaspectratio]{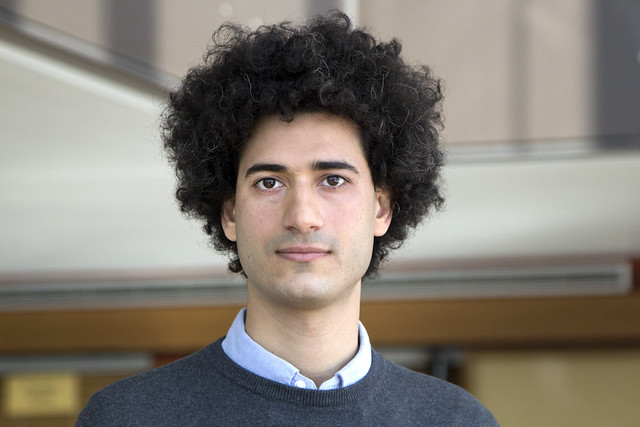}}]{Maani Ghaffari}

received the Ph.D. degree from the Centre for Autonomous Systems (CAS), University of Technology Sydney, NSW, Australia, in 2017. He is currently an Assistant Research Scientist at the Robotics Institute and Department of Naval Architecture and Marine Engineering, University of Michigan, Ann Arbor, MI, USA. His research interests include applied mathematics, robotic perception, machine learning, and planning under uncertainty with applications in robotics and autonomous systems.
\end{IEEEbiography}

\begin{IEEEbiography}[{\includegraphics[width=1in,height=1.25in,clip,trim={0 0.03cm 0 0},keepaspectratio]{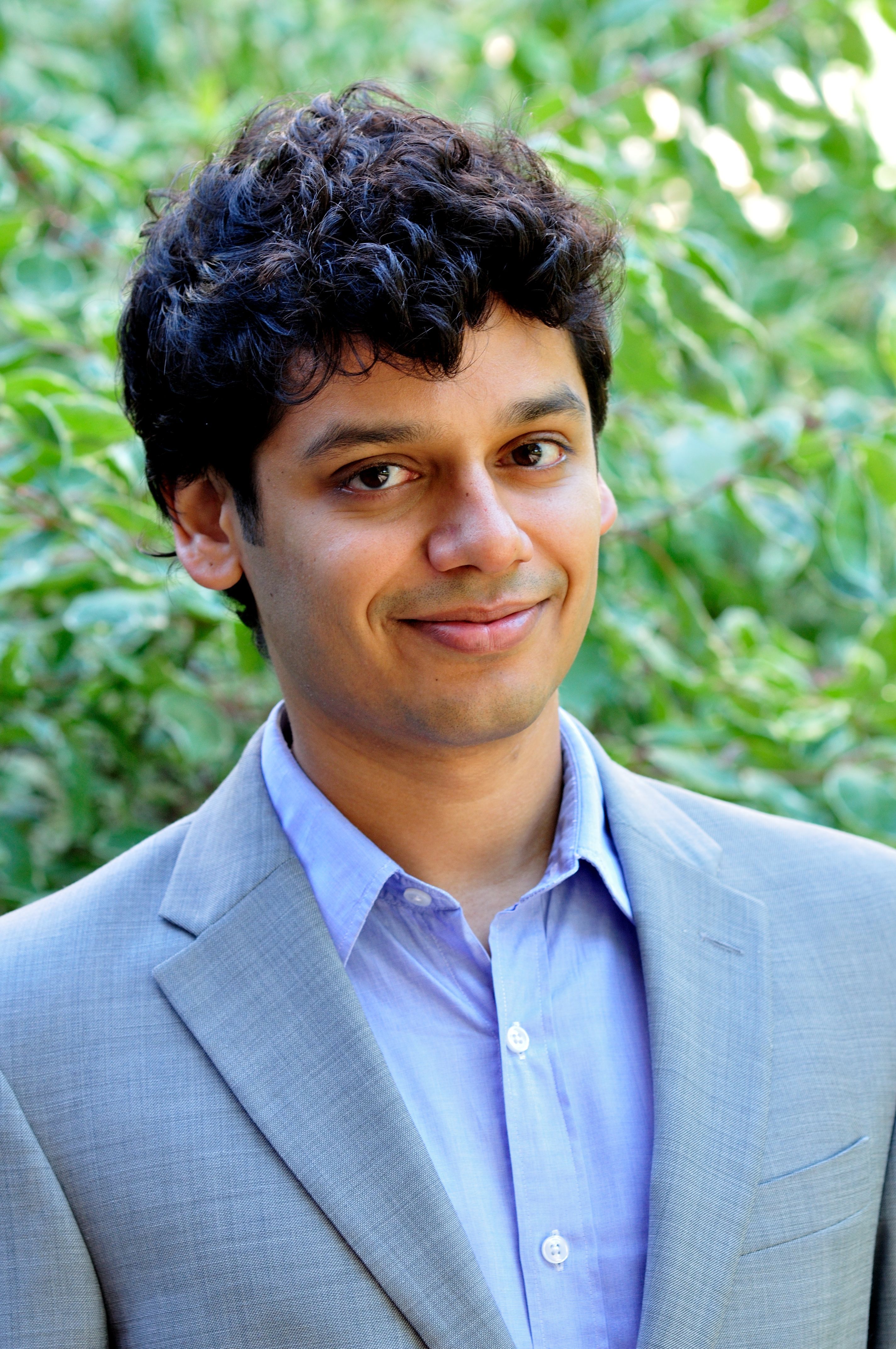}}]{Ram Vasudevan}

received the B.S. degree in electrical engineering and computer sciences, the M.S. degree in electrical engineering, and the Ph.D. degree in electrical engineering from the University of California at Berkeley in 2006, 2009, and 2012, respectively.
He is currently an Assistant Professor in mechanical engineering with the University of Michigan, Ann Arbor, with an appointment in the University of Michigan’s Robotics Program. His research interests include the development and application of optimization and systems theory to quantify and improve human and robot interaction.
\end{IEEEbiography}

\vspace{-1 cm}

\begin{IEEEbiography}[{\includegraphics[width=1in,height=1.25in,clip,trim={1.1cm 0 0.1cm 0},keepaspectratio]{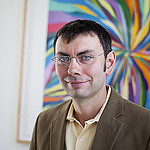}}]{Ryan M. Eustice}

received the B.S. degree in mechanical engineering from Michigan State University, East Lansing, MI, USA, in 1998 and the Ph.D. degree in ocean engineering from the Joint Program between the Massachusetts Institute of Technology (MIT), Cambridge, MA, USA and the Woods Hole Oceanographic Institution (WHOI), Woods Hole, MA, USA, in 2005. Currently, he is the Senior Vice President of Automated Driving at the Toyota Research Institute and a Professor with the Department of Naval Architecture and Marine Engineering, University of Michigan, Ann Arbor, MI, USA, with joint appointments in the Department of Electrical Engineering and Computer Science and in the Department of Mechanical Engineering. His research interests include autonomous navigation and mapping, automated driving, mobile robotics, and autonomous underwater vehicles.

\end{IEEEbiography}

\end{document}